\documentclass{article}

\usepackage{microtype}
\usepackage{graphicx}
\usepackage{subfigure}
\usepackage{amsthm}
\usepackage{amsfonts}
\usepackage{xcolor}
\usepackage{booktabs} 
\usepackage[utf8]{inputenc}
\usepackage[english]{babel}
\usepackage{amsmath}
\usepackage{multicol}
\usepackage{enumitem}
\usepackage{hyperref}

\usepackage{floatrow}
\newfloatcommand{capbtabbox}{table}[][\FBwidth]

\usepackage{natbib}
\usepackage{hyperref}

\def\x{\mathbf{x}}

\usepackage[accepted]{icml2019}

\icmltitlerunning{Concept Whitening for Interpretable Image Recognition}

\begin{document}

\twocolumn[
\icmltitle{Concept Whitening for Interpretable Image Recognition}



\icmlsetsymbol{equal}{*}

\begin{icmlauthorlist}
\icmlauthor{Zhi Chen}{dukecs}
\icmlauthor{Yijie Bei}{dukeece}
\icmlauthor{Cynthia Rudin}{dukecs,dukeece}
\end{icmlauthorlist}

\icmlaffiliation{dukecs}{Department of Computer Science, Duke University, USA}
\icmlaffiliation{dukeece}{Department of Electrical and Computer Engineering, Duke University, USA}

\icmlcorrespondingauthor{Zhi Chen}{zhi.chen1@duke.edu}

\icmlkeywords{Interpretability, Deep learning, Disentangled Representation}

\vskip 0.3in
]



\printAffiliationsAndNotice{}  

\begin{abstract}
What does a neural network encode about a concept as we traverse through the layers? Interpretability in machine learning is undoubtedly important, but the calculations of neural networks are very challenging to understand. Attempts to see inside their hidden layers can either be misleading, unusable, or rely on the latent space to possess properties that it may not have. In this work, rather than attempting to analyze a neural network posthoc, we introduce a mechanism, called \textit{concept whitening} (CW), to alter a given layer of the network to allow us to better understand the computation leading up to that layer. When a concept whitening module is added to a CNN, the axes of the latent space are aligned with known concepts of interest. By experiment, we show that CW can provide us a much clearer understanding for how the network gradually learns concepts over layers. CW is an alternative to a batch normalization layer in that it normalizes, and also decorrelates (whitens) the latent space. CW can be used in any layer of the network without hurting predictive performance.
\end{abstract}

\section{Introduction} \label{sec:introduction}
An important practical challenge that arises with neural networks is the fact that the units within their hidden 
layers are not usually semantically understandable. This is particularly true with computer vision applications, where an expanding body of research has focused centrally on explaining the calculations of neural networks and other black box models. 
Some of the core questions considered in these posthoc analyses of neural networks include: ``What concept does a unit in a hidden layer of a trained neural network represent?''or ``Does this unit in the network represent a concept that a human might understand?'' 

The questions listed above are important, but it is not clear that they would naturally have satisfactory answers when performing posthoc analysis on a pretrained neural network. In fact, there are several reasons why various types of posthoc analyses would not answer these questions. 

Efforts to interpret individual nodes of pretrained neural networks  (e.g. \cite{zhou2018interpreting, zhou2014object}) have shown that some fraction of nodes can be identified to be aligned with some high-level semantic meaning, but these special nodes do not provably contain the network's full information about the concepts. That is, the nodes are not  ``pure,'' and information about the concept could be scattered throughout the network.

Concept-vector methods also \cite{kim2018interpretability,zhou2018interpretable,ghorbani2019towards} have been used to analyze pretrained neural networks. Here, vectors in the latent space are chosen to align with pre-defined or automatically-discovered concepts. While concept-vectors are more promising, they still make the assumption that the latent space of a neural network admits a posthoc analysis of a specific form. In particular, they assume that the latent space places members of each concept in one easy-to-classify portion of latent space. Since the latent space was not explicitly constructed to have this property, there is no reason to believe it holds.

Ideally, we would want a neural network whose latent space \textit{tells us} how it is disentangling concepts, without needing to resort to extra classifiers like concept-vector methods \cite{kim2018interpretability,ghorbani2019towards}, without surveys to humans \cite{zhou2014object}, and without other manipulations that rely on whether the geometry of a latent space serendipitously admits analysis of concepts. Rather than having to rely on assumptions that the latent space admits disentanglement, we would prefer to constrain the latent space directly. We might even wish that the concepts align themselves along the axes of the latent space, so that each point in the latent space has an interpretation in terms of known concepts.

Let us discuss how one would go about imposing such constraints on the latent space. In particular, we introduce the possibility of what we call \textit{concept whitening}. Concept whitening (CW) is a module inserted into a neural network. It constrains the latent space to represent target concepts and also provides a straightforward means to extract them. It does not force the concepts to be learned as an intermediate step, rather \textit{it imposes the latent space to be aligned along the concepts}. 

For instance, let us say that, using CW on a lower layer of the network, the concept ``airplane'' is represented along one axis. By examining the images along this axis, we can find the lower-level characteristic that the network is using to best approximate the complex concept ``airplane,'' which might be white or silver objects with blue backgrounds. In the lower layers of a standard neural network, we cannot necessarily find these characteristics, because the relevant information of ``airplane'' might be spread throughout latent space rather than along an ``airplane'' axis.

By looking at images along the airplane axis at each layer, we see how the network gradually represents airplanes with an increasing level of sophistication and complexity. 

Concept whitening could be used to replace a plain batch normalization step in a CNN backbone, because it combines batch whitening with an extra step involving a rotation matrix. Batch whitening usually provides helpful properties to latent spaces, but our goal requires the whitening to take place with respect to concepts; the use of the rotation matrix to align the concepts with the axes is the key to interpretability through disentangled concepts. Whitening decorrelates and normalizes each axis (i.e., transforms the post-convolution latent space so that the covariance matrix between channels is the identity). 

Exploiting the property that a whitening transformation remains valid after applying arbitrary rotation, the rotation matrix strategically matches the concepts to the axes. 

The concepts used in CW do not need to be the labels in the classification problem, they can be learned from an auxiliary dataset in which concepts are labeled. The concepts do not need to be labeled in the dataset involved in the main classification task (though they could be), and the main classification labels do not need to be available in the auxiliary concept dataset. 

Through qualitative and quantitative experiments, we illustrate how concept whitening applied to the various layers of the neural network illuminates its internal calculations. We verify the interpretability and pureness of concepts in the disentangled latent space. Importantly for practice, we show that by replacing the batch normalization layer in pretrained state-of-the-art models with a CW module, the resulting neural network can achieve accuracy on par with the corresponding original black box neural network on large datasets, and it can do this within one additional epoch of further training. Thus, with fairly minimal effort, one can make a small modification to a neural network architecture (adding a CW module), and in return be able to easily visualize how the network is learning all of the different concepts at any chosen layer. 

CW can show us how a concept is represented at a given layer of the network. What we find is that at lower layers, since a complex concept cannot be represented by the network, it often creates lower-level \textit{abstract concepts}. For example, an airplane at an early layer is represented by an abstract concept defined by white or gray objects on a blue background. A bed is represented by an abstract concept that seems to be characterized by warm colors (orange, yellow). In that sense, the CW layer can help us to discover new concepts that can be formally defined and built on, if desired.

\section{Related work}
\label{sec:related_work}
There are several large and rapidly expanding bodies of relevant literature.

\textbf{Interpretability and explainability of neural networks:}\\
There have been two schools of thought on improving the interpretability of neural networks: (1) learning an inherently \textit{interpretable} model \cite{Rudin2019}; (2) providing post-hoc \textit{explanations} for an exist neural network. CW falls within the first type, though it only enlightens what the network is doing, rather than providing a full understanding of the network's computations. To provide a full explanation of each computation would lead to more constraints and thus a loss in flexibility, whereas CW allows more flexibility in exchange for more general types of explanations. The vast majority of current works on neural networks are of the second type, explainability. A problem with the terminology is that ``explanation'' methods are often summary statistics of performance (e.g., local approximations, general trends on node activation) rather than actual explanations of the model's calculations. For instance, if a node is found to activate when a certain concept is present in an image, it does not mean that all information (or even the majority of information) about this concept is involved with that particular node.

Saliency-based methods are the most common form of post-hoc explanations for neural networks \cite{zeiler2014visualizing, simonyan2013deep,smilkov2017smoothgrad,selvaraju2017grad}. These methods assign importance weights to each pixel of the input image to show the importance of each pixel to the image's predicted class. Saliency maps are problematic for well-known reasons: they often provide highlighting of edges in images, regardless of the class. Thus, very similar explanations are given for multiple classes, and often none of them are useful explanations \cite{Rudin2019}. 
Saliency methods can be unreliable and fragile \cite{adebayo2018sanity}.

Other work provides explanations of how the network's latent features operate. Some measure the alignment of an individual internal unit, or a filter of a trained neural network, to a predefined concept and find some units have relatively strong alignment to that concept \cite{zhou2018interpreting,zhou2014object}. While some units (i.e., filters) may align nicely with pre-defined concepts, the concept can be represented diffusely through many units (the concept representation by individual nodes is impure); this is because the network was not trained to have concepts expressed purely through individual nodes. To address this weakness, several concept-based post-hoc explanation approaches have recently been proposed that do not rely on the concept aligning with individual units \cite{kim2018interpretability,zhou2018interpretable,ghorbani2019towards,yeh2019concept}. Instead of analyzing individual units, these methods try to learn a linear combination of them to represent a predefined concept \cite{kim2018interpretability} or to automatically discover concepts by clustering patches and defining the clusters as new concepts \cite{ghorbani2019towards}. Although these methods are promising, they are based on assumptions of the latent space that may not hold. For instance, these methods assume that a classifier (usually a linear classifier) exists on the latent space such that the concept is correctly classified. Since the network was not trained so that this assumption holds, it may not hold. More importantly, since the latent space is not shaped explicitly to handle this kind of concept-based explanation, unit vectors (directions) in the latent space may not represent concepts purely. We will give an example in the next section to show why latent spaces built without constraints may not achieve concept separation.

CW avoids these problems because it shapes the latent space through training. In that sense, CW is closer to work on inherently interpretable neural networks, though its use-case is in the spirit of concept vectors, in that it is useful for providing important directions in the latent space.

There are emerging works trying to build inherently interpretable neural networks. Like CW, they alter the network structure to encourage different forms of interpretability. For example, neural networks have been designed to perform case-based reasoning \cite{chen2019looks, li2018deep}, to incorporate logical or grammatical structures \cite{li2017aognets, granmo2019convolutional,wu2019towards}, to do classification based on hard attention \cite{mnih2014recurrent, ba2014multiple,sermanet2014attention,elsayed2019saccader}, or to do image recognition by decomposing the components of images \cite{saralajew2019classification}. These models all have different forms of interpretability than we consider (understanding how the latent spaces of each layer can align with a known set of concepts). Other work also develops inherently interpretable deep learning methods that can reason based on concepts, but are different from our work in terms of field of application \cite{bouchacourt2019educe}, types of concepts \cite{zhang2018interpretable,zhang2018unsupervised} and ways to obtain concepts \cite{adel2018discovering}.

In the field of deep generative models, many works have been proposed to make the latent space more interpretable by forcing disentanglement. However, works such as InfoGAN \cite{chen2016infogan} and $\beta$-VAE \cite{higgins2017beta}, all use heuristic interpretability losses like mutual information, while in CW we have actual concepts that we use to align the latent space.

\textbf{Whitening and orthogonality:}
Whitening is a linear transformation that transforms the covariance matrix of random input vectors to be the identity matrix. It is a classical preprocessing step in data science. In the realm of deep learning, batch normalization \cite{ioffe2015batch}, which is widely used in many state-of-the-art neural network architectures, retains the standardization part of whitening but not the decorrelation. Earlier attempts whiten by periodically estimating the whitening matrix \cite{desjardins2015natural,luo2017learning}, which leads to instability in training. Other methods perform whitening by adding a decorrelation loss \cite{cogswell2015reducing}. 
A whitening module for ZCA has been developed that leverages the fact that SVD is differentiable, and supports backpropagation  \cite{huang2018decorrelated, huang2019iterative}.
Similarly, others have developed a differentiable whitening block based on Cholesky whitening \cite{siarohin2018whitening}. The whitening part of our CW module borrows techiques from IterNorm \cite{huang2019iterative} because it is differentiable and accelerated. CW is different from previous methods because its whitening matrix \textit{is multiplied by an orthogonal matrix} and \textit{maximizes the activation of known concepts along the latent space axes}.

In the field of deep learning, many earlier works that incorporate orthogonality constraints are targeted for RNNs \cite{vorontsov2017orthogonality,mhammedi2017efficient,wisdom2016full}, since orthogonality could help avoid vanishing gradients or exploding gradients in RNNs. Other work explores ways to learn orthogonal weights or representations for all types of neural networks (not just RNNs) \cite{harandi2016generalized, huang2018orthogonal,lezcano2019cheap,lezama2018ole}. For example, some work \cite{lezama2018ole} uses special loss functions to force orthogonality. The optimization algorithms used in the above methods are all different from ours. For CW, we optimize the orthogonal matrix by Cayley-transform-based curvilinear search algorithms \cite{wen2013feasible}. While some deep learning methods also use a Cayley transform \cite{vorontsov2017orthogonality}, they do it with a fixed learning rate that does not work effectively in our setting. More importantly, the goal of doing optimization with orthogonality constraints in all these works are completely different from ours. None of them try to align columns of the orthogonal matrix with any type of concept.

\section{Methodology}
\label{sec:methodology}
Suppose $\mathbf{x}_1,\mathbf{x}_2,...,\mathbf{x}_n\in \mathcal{X}$ are samples in our dataset and $y_1,y_2,..y_n\in\mathcal{Y}$ are their labels. From the latent space $\mathcal{Z}$ defined by a hidden layer, a DNN classifier $f:\mathcal{X}\rightarrow\mathcal{Y}$ can be divided into two parts, a feature extractor
$\mathbf{\Phi}:\mathcal{X}\rightarrow\mathcal{Z}$, with parameters $\theta$, and a classifier $g:\mathcal{Z}\rightarrow\mathcal{Y}$, parameterized by $\omega$. Then $\mathbf{z}=\mathbf{\Phi}(\mathbf{x};\theta)$ is the latent representation of the input $\mathbf{x}$ and $f(\mathbf{x})=g(\mathbf{\Phi}(\mathbf{x};\theta);\omega)$ is the predicted label.
Suppose we are interested in $k$ concepts called $c_1,c_2,...c_k$. We can then pre-define $k$ auxiliary datasets $\mathbf{X}_{c_1},\mathbf{X}_{c_2}...,\mathbf{X}_{c_k}$ such that samples in $\mathbf{X}_{c_j}$ are the most representative samples of concept $c_j$. Our goal is to learn $\mathbf{\Phi}$ and $g$ simultaneously, such that (a) the classifier $g(\mathbf{\Phi}(\cdot;\theta);\omega)$ can predict the label accurately; (b) the $j^{th}$ dimension $z_j$ of the latent representation $\mathbf{z}$ aligns with concept $c_j$. In other words, samples in $\mathbf{X}_{c_j}$ should have larger values of $z_j$ than other samples. Conversely, samples not in $\mathbf{X}_{c_j}$ should have smaller values of $z_j$.

\subsection{Standard Neural Networks May Not Achieve Concept Separation} \label{sec:whynotnn}
Some posthoc explanation methods have looked at unit vectors in the direction of data where a concept is exhibited; this is done to measure how different concepts contribute to a classification task \cite{zhou2018interpretable}. Other methods consider directional derivatives towards data exhibiting the concept \cite{kim2018interpretability}, for the same reason. There are important reasons why these types of approaches may not work.

\begin{figure}[htbp]
    \centering
    \includegraphics[width=80mm]{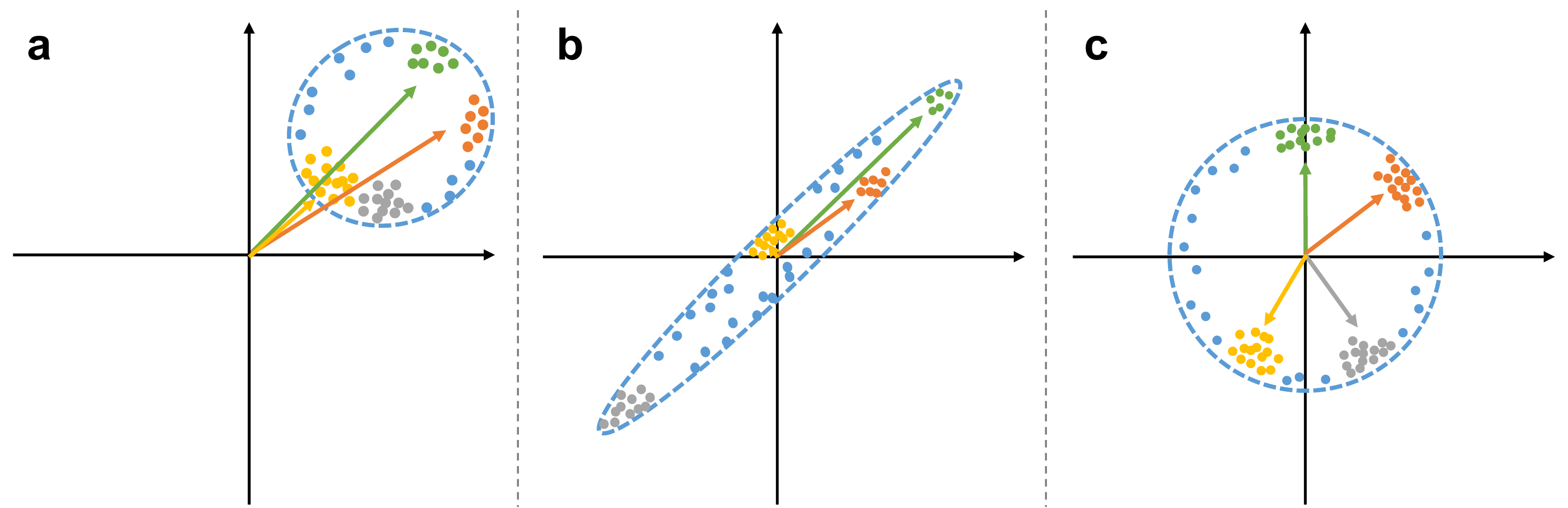}
    \caption{Possible data distributions in the latent space. \textbf{a}, the data are not mean centered; \textbf{b} the data are standardized but not decorrelated; \textbf{c} the data are whitened. In both \textbf{a} and \textbf{b},  unit vectors are not valid for representing concepts. \label{fig:demo}}
\end{figure}

First, suppose the latent space is not mean-centered. This alone could cause problems for posthoc methods that compute directions towards concepts. Consider, for instance, a case where all points in the latent space are far from the origin. In that case, \textit{all} concept directions point towards the same part of the space: the part where the data lies (see Figure \ref{fig:demo}(a)).
This situation might be fairly easy to solve since the users can just analyze the latent space of a batch normalization layer or add a bias term. But then other problems could arise.

Even if the latent space is mean-centered and standardized, the latent space of standard neural networks may not separate concepts. Consider, for instance, an elongated latent space similar to that illustrated in Figure \ref{fig:demo}(b), by the green and orange clusters. Here, two unit vectors pointing to different groups of data (perhaps exhibiting two separate concepts) may have a large inner product, suggesting that they may be part of the same concept, when in fact, they may be not be similar at all, and may not even lie in the same part of the latent space. Thus, even if the latent space is standardized, multiple unrelated concepts can still appear similar because, from the origin, their centers point towards the same general direction. For the same reason, taking derivatives towards the parts of the space where various concepts tend to appear may yield similar derivatives for very different concepts.

For the above reasons, a latent space in which unit vectors can effectively represent different concepts should have small inter-concept similarity (as illustrated in Figure \ref{fig:demo}(c)). That is, samples of different concepts should be near orthogonal in the latent space. In addition, for better concept separation, the ratio between inter-concept similarity and intra-concept similarity should be as small as possible.

The CW module we introduce in this work can make the latent space \textit{mean-centered and decorrelated}. This module can align predefined concepts in orthogonal directions. More details of the proposed module will be discussed in Section \ref{sec:CW} and Section \ref{sec:detail}. The experimental results in Section \ref{sec:inter_inner} compare the inter-concept and intra-concept similarity of the latent space of standard NNs with and without the proposed CW module. The results validate that the previously mentioned problems of standard neural networks do exist, and that the proposed method successfully avoids these problems.

\subsection{Concept Whitening Module}
\label{sec:CW}
Let $\mathbf{Z}_{d \times n}$ be the latent representation matrix of $n$ samples, in which each column $\mathbf{z}_i\in \mathbb{R}^d$ contains the latent features of the $i^{th}$ sample. Our Concept Whitening module (CW) consists of two parts, whitening and orthogonal transformation. The whitening transformation $\psi$ decorrelates and standardizes the data by
\begin{equation}
    \psi(\mathbf{Z}) = \mathbf{W}(\mathbf{Z}-\mathbf{\mu}{\mathbf{1}_{n\times 1}}^T)
\end{equation}
where $\mathbf{\mu}=\frac{1}{n}\sum_{i=1}^{n}\mathbf{z}_i$ is the sample mean and $\mathbf{W}_{d \times d}$ is the whitening matrix that obeys $\mathbf{W}^T\mathbf{W}=\mathbf{\Sigma}^{-1}$. Here, $\mathbf{\Sigma}_{d \times d}=\frac{1}{n}(\mathbf{Z}-\mathbf{\mu}\mathbf{1}^T)(\mathbf{Z}-\mathbf{\mu}\mathbf{1}^T)^T$ is the covariance matrix. The whitening matrix $\mathbf{W}$ is not unique and can be calculated in many ways such as ZCA whitening and Cholesky decomposition. Another important property of the whitening matrix is that it is rotation free; suppose $\mathbf{Q}$ is an orthogonal matrix, then
\begin{equation}
    \mathbf{W'}=\mathbf{Q^TW}
\end{equation}
is also a valid whitening matrix. In our module, after whitening the latent space to endow it with the properties discussed above, we still need to rotate the samples in their latent space such that the data from concept $c_j$, namely $\mathbf{X}_{c_j}$, are highly activated on the $j^{th}$ axis. Specifically, we need to find an orthogonal matrix $\mathbf{Q}_{d \times d}$ whose column $\mathbf{q}_j$ is the $j^{th}$ axis, by optimizing the following objective:
\begin{equation}
    \begin{aligned}
        \max_{\mathbf{q}_1,\mathbf{q}_2,...,\mathbf{q}_k}  &\sum_{j=1}^{k}\frac{1}{n_j}\mathbf{q}_j^T\psi(\mathbf{Z}_{c_j})\mathbf{1}_{n_j\times 1} \\
        s.t.\ &\mathbf{Q}^T\mathbf{Q}=\mathbf{I}_d
    \end{aligned}
\end{equation}
where $\mathbf{Z}_{c_j}$ is a $d \times n_j$ matrix denoting the latent representation of $\mathbf{X}_{c_j}$ and $c_1,c_2, ...,c_k$ are concepts of interest. An optimization problem with an orthogonality constraint like this can be solved by gradient-based approaches on the Stiefel manifold (e.g., the method of \cite{wen2013feasible}).

This whole procedure constitutes CW, and can be done for any given layer of a neural network as part of the training of the network. The forward pass of the CW module, which makes predictions, is summarized in Algorithm \ref{alg:forward}.

\begin{algorithm*}[ht]
 \caption{Forward Pass of CW Module}
 \begin{tabbing}
 xxx \= xx \= xx \= xx \= xx \= xx \kill
 1: \> \textbf{Input}: mini-batch input $\mathbf{Z} \in \mathbb{R}^{d\times m}$ \\
 2: \> \textbf{Optimization Variables:} orthogonal matrix $\mathbf{Q}\in \mathbb{R}^{d\times d}$ (learned in Algorithm 2)\\
 3: \> \textbf{Output:} whitened representation $\mathbf{\hat Z} \in \mathbb{R}^{d\times m}$ \\
 4: \> calculate batch mean: $\mu = \frac{1}{m} \mathbf{Z}\cdot\mathbf{1}$, and center the activation: $\mathbf{Z_C} = \mathbf{Z} - \mu\cdot\mathbf{1}^T$ \\
 5: \> calculate ZCA-whitening matrix $\mathbf{W}$, for details see Algorithm 1 of \cite{huang2019iterative}  \\
 6: \> calculate the whitened representation: $\mathbf{\hat Z} = \mathbf{Q}^T\mathbf{W}\mathbf{Z_C}$.
 \end{tabbing}
 \label{alg:forward}
\end{algorithm*}

\begin{algorithm*}[ht]
 \caption{Alternating Optimization Algorithm for Training}
 \begin{tabbing}
 xxx \= xx \= xx \= xx \= xx \= xx \kill
 1: \> \textbf{Input}: main objective dataset $\mathcal{D} = \{\mathbf{x}_i,y_i\}_{i=1}^{n} $, concept datasets $\mathbf{X}_{c_1},\mathbf{X}_{c_2}...,\mathbf{X}_{c_k}$ \\
 2: \> \textbf{Optimization Variables}: $\theta$, $\omega$, $\mathbf{W}$, $\mu$, $\mathbf{Q}$, whose definitions are in Section \ref{sec:CW} \\
 3: \> \textbf{Parameters}: $\beta$, $\eta$\\
 4: \> \textbf{for} $t$ = 1 to $T$ \textbf{do} \\
 5: \> \> randomly sample a mini-batch  $\{\mathbf{x}_i,y_i\}_{i=1}^{m}$ from $\mathcal{D}$ \\
 6: \> \> do one step of SGD w.r.t. $\theta$ and $\omega$ on the loss $\frac{1}{m}\sum_{i=1}^{m}\ell(g(\mathbf{Q}^T\psi(\mathbf{\Phi}(\mathbf{x}_i;\theta);\mathbf{W},\mathbf{\mu});\omega),y_i)$ \\
 7: \> \> update $\mathbf{W}$ and $\mu$ by exponential moving average \\
 8: \> \> \textbf{if} \ $t\hspace*{-3pt}\mod 20 = 0$ \textbf{then} \\
 9: \> \> \> sample mini-batches $\{\mathbf{x}_i^{(c_1)}\}_{i=1}^{m},\{\mathbf{x}_i^{(c_2)}\}_{i=1}^{m},...,\{\mathbf{x}_i^{(c_k)}\}_{i=1}^{m}$ from $\mathbf{X}_{c_1},\mathbf{X}_{c_2},...,\mathbf{X}_{c_k}$ \\
 10: \> \> \> calculate $\mathbf{G}=\nabla_{\mathbf{Q}}$, with columns $\mathbf{g}_j=-\frac{1}{m}\sum_{i=1}^{m}\psi(\mathbf{\Phi}(\mathbf{x}_{i}^{(c_j)};\theta);\mathbf{W},\mathbf{\mu})$ when $1\leq j \leq k$, else $\mathbf{g}_j=\mathbf{0}$ \\ 
 11: \> \> \> calculate the exponential moving average of $\mathbf{G}$: $\mathbf{G'} =  \beta \mathbf{G'} + (1-\beta) \mathbf{G}$ \\
 12: \> \> \> obtain learning rate $\eta$ by curvilinear search, for details see Algorithm 1 of \cite{wen2013feasible}  \\
 13: \> \> \> update $\mathbf{Q}$ by Cayley transform:  $\mathbf{Q}\leftarrow(I+\frac{\eta}{2}(\mathbf{G'}\mathbf{Q}^T-\mathbf{Q}\mathbf{G'}^T))^{-1}(I-\frac{\eta}{2}(\mathbf{G'}\mathbf{Q}^T-\mathbf{Q}\mathbf{G'}^T))\mathbf{Q}$
 \end{tabbing}
 \label{alg:twostep}
\end{algorithm*}

\subsection{Optimization and Implementation Detail} \label{sec:detail}
Whitening has not (to our knowledge) been previously applied to align the latent space to concepts. In the past, whitening has been used to speed up back-propagation. The specific whitening problem for speeding up back-propagation is different from that for concept alignment--the rotation matrix is not present in other work on whitening, nor is the notion of a concept--however, we can leverage some of the optimization tools used in that work on whitening \cite{huang2019iterative,huang2018orthogonal,siarohin2018whitening}. Specifically, we adapt ideas underlying the IterNorm algorithm \cite{huang2019iterative}, which employs Newton's iterations to approximate ZCA whitening, to the problem studied here. Let us now describe how this is done.

The whitening matrix in ZCA is
\begin{equation}
    \mathbf{W} = \mathbf{D}\mathbf{\Lambda}^{-\frac{1}{2}}\mathbf{D}^T
\end{equation}
where $\mathbf{\Lambda}_{d \times d}$ and $\mathbf{D}_{d \times d}$ are the eigenvalue diagonal matrix and eigenvector matrix given by the eigenvalue decomposition of the covariance matrix,  $\mathbf{\Sigma}=\mathbf{D}\mathbf{\Lambda}\mathbf{D}^T$. Like other normalization methods, we calculate a $\mathbf{\mu}$ and $\mathbf{W}$ for each mini-batch of data, and average them together to form the model used in testing.

As mentioned in Section \ref{sec:CW}, the challenging part for CW is that we also need to learn an orthogonal matrix by solving an optimization problem. To do this, we will optimize the objective while strictly maintaining the matrix to be orthogonal by performing gradient descent with a curvilinear search on the Stiefel manifold \cite{wen2013feasible} and adjust it to deal with mini-batch data.

\textbf{The two step alternating optimization}: During training, our procedure must handle two types of data: data for calculating the main objective and the data representing the predefined concepts. The model is optimized by alternating optimization: the mini-batches of the main dataset and the auxiliary concept dataset are fed to the network, and the following two objectives are optimized in turns. The first objective is the main objective (usually related to classification accuracy):
\begin{equation}
    \min_{\theta, \omega, W, \mu} \frac{1}{n}\sum_{i=1}^{n}\ell(g(\mathbf{Q}^T\psi(\mathbf{\Phi}(\mathbf{x}_i;\theta);\mathbf{W},\mathbf{\mu});\omega),y_i)
\end{equation}
where $\mathbf{\Phi}$ and $g$ are layers before and after the CW module parameterized by $\theta$ and $\omega$ respectively. $\psi$ is a whitening transformation parameterized by sample mean $\mathbf{\mu}$ and whitening matrix $\mathbf{W}$. $\mathbf{Q}$ is the orthogonal matrix. The combination $\mathbf{Q}^T\psi$ forms the CW module (which is also a valid whitening transformation). $\ell$ is any differentiable loss. We use cross-entropy loss for $\ell$ in our implementation to do classification, since it is the most commonly used. The second objective is the concept alignment loss:
\begin{equation}
    \begin{aligned}
        \max_{\mathbf{q}_1,\mathbf{q}_2,...,\mathbf{q}_k}  &\sum_{j=1}^{k}\frac{1}{n_j}\sum_{x_i^{(c_j)}\in X_{c_j}}\mathbf{q}_j^T\psi(\mathbf{\Phi}(\mathbf{x}_{i}^{(c_j)};\theta);\mathbf{W},\mathbf{\mu}) \\
        s.t.\ &\mathbf{Q}^T\mathbf{Q}=\mathbf{I}_d.
    \end{aligned}
\end{equation}
The orthogonal matrix $\mathbf{Q}$ is fixed when training for the main objective and the other parameters are fixed when training for $\mathbf{Q}$. The optimization problem is a linear programming problem with quadratic constraints (LPQC) which is generally NP-hard. Since directly solving for the optimal solution is intractable, we optimize it by gradient methods on the Stiefel manifold. 
At each step $t$, in which the second objective is handled, the orthogonal matrix $\mathbf{Q}$ is updated by Cayley transform
$$\mathbf{Q}^{(t+1)}=\left(I+\frac{\eta}{2}\mathbf{A}\right)^{-1}\left(I-\frac{\eta}{2}\mathbf{A}\right)\mathbf{Q}^{(t)}$$
where $\mathbf{A}=\mathbf{G}(\mathbf{Q}^{(t)})^T-\mathbf{Q}^{(t)}\mathbf{G}^T$ is a skew-symmetric matrix, $\mathbf{G}$ is the gradient of the loss function and $\eta$ is the learning rate. The optimization procedure is accelerated by curvilinear search on the learning rate at each step \cite{wen2013feasible}. Note that, in the Cayley transform, the stationary points are reached when $\mathbf{A}=\mathbf{0}$, which has multiple solutions. Since the solutions are in high-dimensional space, these stationary points are very likely to be saddle points, which can be avoided by SGD. Therefore, we use the stochastic gradient calculated by a mini-batch of samples to replace $\mathbf{G}$ at each step. To accelerate and stabilize the stochastic gradient, we also apply momentum to it during implementation. Algorithm \ref{alg:twostep} provides details for the two-step alternating optimization.

\textbf{Dealing with the convolution outputs:} In the previous description of our optimization algorithm, we assume that the activations in the latent space form a vector. However, in CNNs, the output of the layer is a tensor instead of a vector. In CNNs, a feature map (a channel within one layer, created by a convolution of one filter) contains the information of how activated a part of the image is by a single filter, which may be a detector for a specific concept. Let us reshape the feature map into a vector, where each element of the vector represents how much one part of the image is activated by the filter. Thus, if the feature map for one filter is $h \times w$ then a vector of length $hw$ contains the activation information for that filter around the whole feature map. We do this reshaping procedure for each filter, which reshapes the output of a convolution layer $Z_{h\times w \times d \times n}$ into a matrix $Z_{d\times(hwn)}$, where $d$ is the number of channels. We then perform CW on the reshaped matrix. After doing this, the resulting matrix is still size $d\times(hwn)$. If we reshape this matrix back to its original size as a tensor, one feature map of the tensor now (after training) represents whether a meaningful concept is detected at each location in the image for that layer. Note that, now the output of a filter is a feature map which is a $h\times w$ matrix but the concept activation score we used in the optimization problem is a scalar. Therefore, we need to get an activation value from the feature map. There are multiple ways to do this. We try the following calculations to define activation based on the feature map: (a) mean of all feature map values; (b) max of all feature map values; (c) mean of all positive feature map values; (d) mean of down-sampled feature map obtained by max pooling.  We use (d) in our experiments since it is good at capturing both high-level and low-level concepts. Detailed analysis and experiments about the choice of different activation calculations are discussed in Supplementary Information \ref{sec:activation}.

\textbf{Warm start with pretrained models:} Let us discuss some aspects of practical implementation. The CW module can substitute for other normalization modules such as BatchNorm in an hidden layer of the CNN. Therefore, one can use the weights of a pretrained model as a warm start. To do this, we might leverage a pretrained model (for the same main objective) that does not use CW, and replace a BatchNorm layer in that network with a CW layer. The model usually converges in one epoch (one pass over the data) if a pretrained model is used.

Note that CW does strictly more work than BatchNorm. CW alone will achieve desirable outcomes of using BatchNorm, therefore, there is no need to use BatchNorm when CW is in place. 

\textbf{Computational Efficiency:} The CW module involves two iterative optimization steps: one for whitening normalization and one for concept alignment. The efficiency of iterative whitening normalization is justified experimentally in \cite{huang2019iterative}; the concept alignment optimization is performed only every 20 batches, usually costing less than 20 matrix multiplications and 10 matrix inversions, which do not notably hurt the speed of training. Indeed, our experiments show that there is no significant training speed slowdown using CW compared to using vanilla BN.

\section{Experiments}
\label{sec:experiment}
In this section, we first show that after replacing one batch norm (BN) layer with our CW module, the accuracy of image recognition is still on par with the original model (\ref{sec:acc}). After that, we visualize the concept basis we learn and show that the axes are aligned with the concepts assigned to them. Specifically, we display the images that are most activated along a single axis (\ref{sec:top10}); we then show how two axes interact with each other (\ref{sec:2drep}); and we further show how the same concept evolves in different layers (\ref{sec:trajectory}), where we have replaced one layer at a time. Then we validate the problems standard neural network mentioned in Section \ref{sec:whynotnn} through experiments and show that CW can solve these problems (\ref{sec:inter_inner}). Moreover, we quantitatively measure the interpretability of our concept axes and compare with other concept-based neural network methods (\ref{sec:quantitative_eval}). We also show how we can use the learned representation to measure the contributions of the concepts (\ref{sec:concept_importance}). Finally, we show the practicality of the CW module through a case study of skin lesion diagnosis (\ref{sec:isic}).

\subsection{Main Objective Accuracy}
\label{sec:acc}
We evaluate the image recognition accuracy of the CNNs before and after adding a CW module. We show that simply replacing a BN module with a CW module and training for a single epoch leads to similar main objective performance. Specifically, after replacing the BN module with the CW module, we trained popular CNN architectures including VGG16+BN \cite{simonyan2014very}, ResNet with 18 layers and 50 layers \cite{he2016deep} and DenseNet161 \cite{huang2017densely} on the Places365  \cite{zhou2017places} dataset.  The auxiliary concept dataset we used is MS COCO \cite{lin2014microsoft}. Each annotation, e.g., ``person'' in MS COCO, was used as one concept, and we selected all the images with this annotation (images having ``person'' in it), cropped them using bounding boxes and used the cropped images as the data representing the concept. The concept bank has 80 different concepts corresponding to 80 annotations in MS COCO. In order to limit the total time of the training process, we used pretrained models for the popular CNN architectures (discussed above) and fine-tuned these models after BN was replaced with CW. 

Table \ref{fig:acc_places} shows the average test accuracy on the validation set of Places365 over 5 runs. We randomly selected 3 concepts from the concept bank to learn using CW for each run, and used the average of them to measure accuracy. We repeated this, applying CW to different layers and reported the average accuracy among the layers. The accuracy does not change much when CW is applied to the different layers and trained on different number of concepts, as shown in Supplementary Information \ref{sec:acc_sensitivity}.  
\begin{table}[htbp]
    \centering
    \begin{tabular}{lcc c cc}
     \hline
     & \multicolumn{2}{c}{Top-1 acc.} && \multicolumn{2}{c}{Top-5 acc.} \\
     \cline{2-3} \cline{5-6}
     & Original & +CW && Original & +CW \\ 
     \hline
     VGG16-BN & 53.6 & 53.3 && 84.2 & 83.8\\ 
     ResNet18 & 54.5 & 53.9 && 84.6 & 84.2\\ 
     ResNet50 & 54.7 & 54.9 && 85.1 &  85.2\\ 
     DenseNet161 & 55.3 & 55.5 && 85.2 &  85.6\\ 
     \hline
    \end{tabular}
    \caption{Top-1 and top-5 test accuracy on Places365 dataset. Our results show that CW does not hurt performance.}
    \label{fig:acc_places}
\end{table}

Because we have leveraged a pretrained model, when training with CW, we conduct only one additional epoch of training (one pass over the dataset) for each run. 
As shown in Table \ref{fig:acc_places}, the performance of these models using the CW module is on par with the original model: the difference is within $1\%$ with respect to top-1 and top-5 accuracy. This means in practice, \textit{if a pretrained model (using BN) exists, one can simply replace the BN module with a CW module and train it for one epoch, in which case, the pretrained black-box model can be turned into a more interpretable model that is approximately equally accurate.}

\subsection{Visualizing the Concept Basis} \label{sec:visualize_basis}
In order to demonstrate the interpretability benefits of models equipped with a CW module, we visualize the concept basis in the CW module and validate that the axes are aligned with their assigned concepts. In detail, (a) we check the most activated images on these axes; (b) we look at how images are distributed in a 2D slice of the latent space; (c) we show how realizations of the same concept change if we apply CW on different layers. All experiments in \ref{sec:visualize_basis} were done on ResNet18 equipped with CW trained on Places365 and three simultanous MS COCO concepts.

\subsubsection{Top-10 Activated Images}
\label{sec:top10}
\begin{figure}[htbp]
    \centering
    \includegraphics[width=80mm]{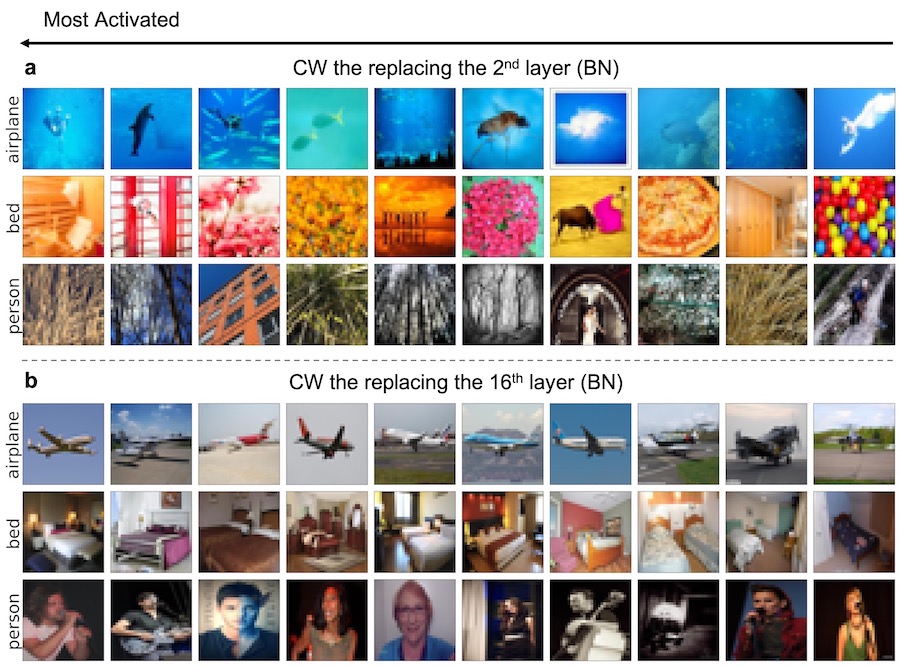}
    \caption{Top-10 Image activated on axes representing different concepts. \textbf{a}, results when the $2^{nd}$ layer (BN) is replaced by CW; \textbf{b}, results when the $16^{th}$ layer (BN) is replaced by CW.\label{fig:top10}}
\end{figure}

We sort all validation samples by their activation values (discussed in Section \ref{sec:detail}) to show how much they are related to the concept. 
Figure \ref{fig:top10} shows the images that have the top-10 largest activations along three different concepts' axes. Note that all these concepts are trained together using one CW module. 

From Figure \ref{fig:top10}(b), we can see that all of the top activated images have the same semantic meaning when the CW module is located at a higher layer (i.e., the 16th layer). Figure \ref{fig:top10}(a) shows that when the CW module is applied to a lower layer (i.e., the 2nd layer), it tends to capture low level information such as color or texture characteristic of these concepts. For instance, the top activated images on the ``airplane'' axis generally has a blue background with a white or gray object in the middle. It is reasonable that the lower layer CW module cannot extract complete information about high-level concepts such as ``airplane'' since the model complexity of the first two layers is limited.

In that sense, the CW layer has \textit{discovered} lower-level characteristics of a more complex concept; namely it has discovered that the blue images with white objects are primitive characteristics that can approximate the ``airplane'' concept. Similarly, the network seems to have discovered that the appearance of warm colors is a lower-level characteristic of the ``bedroom'' concept, and that a dark background with vertical light streaks is a characteristic of the ``person'' concept. 

Interestingly, when different definitions of activation are used (namely the options discussed in Section \ref{sec:detail}), the characteristics discovered by the network often look different. Some of these are shown in Supplementary Information \ref{sec:top10_act}. 

Moreover, similar visualizations show that CW can deal with various types of concepts. Top activated images on more concepts, including concepts defined as objects and concepts defined as general characteristics, can be found in Supplementary Information \ref{sec:more_concepts}. Top activated images visualized with empirical receptive fields can be found in Supplementary Information \ref{sec:receptive_fields}.

\subsubsection{2D-representation space visualization} \label{sec:2drep}

\begin{figure*}[t]
\centering
\includegraphics[width=170mm]{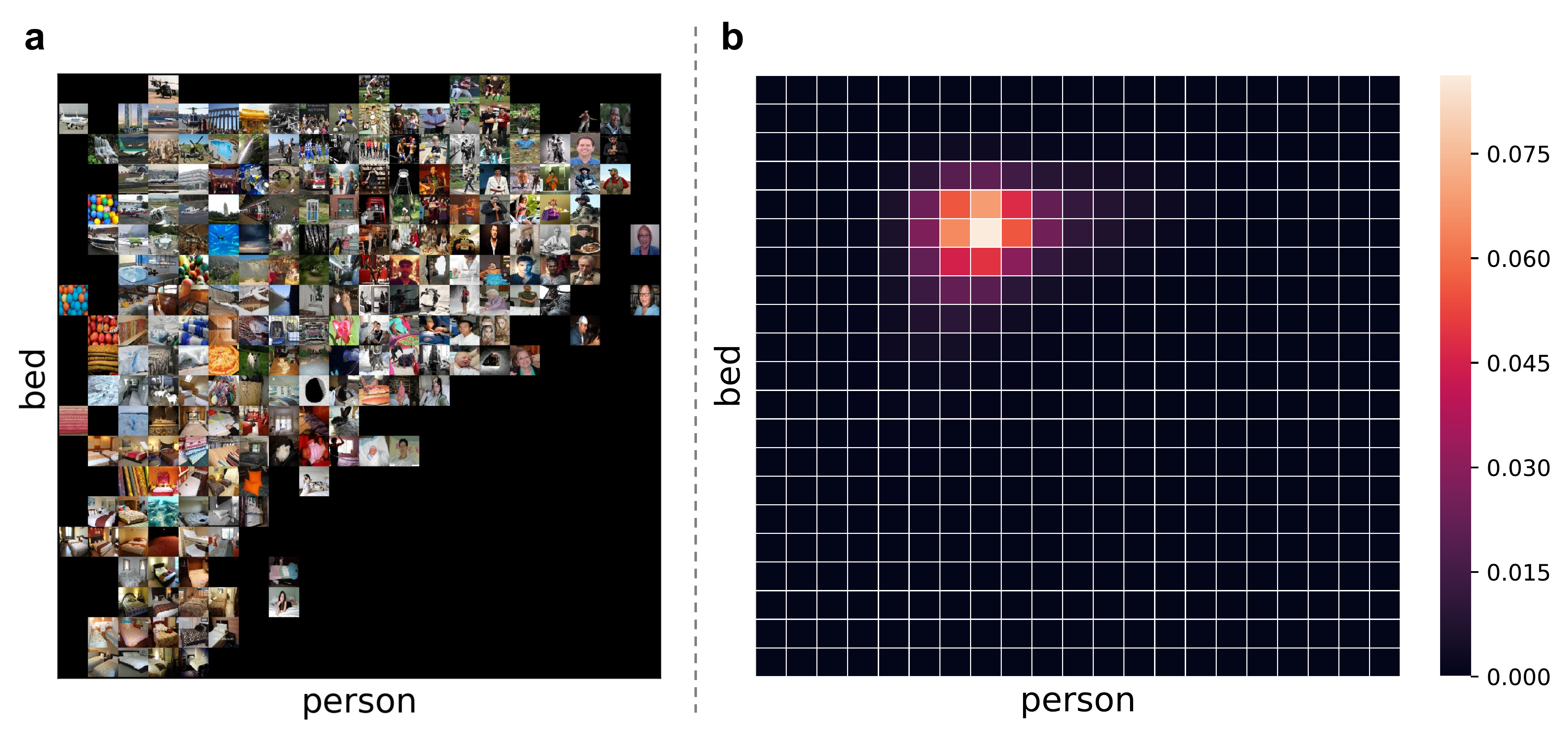}
\caption{Joint distribution of the bed-person subspace. The bounding box given by projected values in the subspace is evenly divided into $20\times20$ blocks. \textbf{a}, Plotting a random test image fall into each block; \textbf{b}, Density map of test image representation\label{fig:2drep}}
\end{figure*}

Let us consider whether joint information about different concepts is captured by the latent space of CW. To investigate how the data are distributed in the new latent space, we pick a 2D slice of the latent space, which means we select two axes $\mathbf{q}_i$ and $\mathbf{q}_j$ and look at the subspace they form.

The data's joint distribution on the two axes is shown in Figure \ref{fig:2drep}. To visualize the joint distribution, we first compute the activations of all validation data on the two axes, then divide the latent space into a $50\times50$ grid of blocks, where the maximum and minimum activation value are the top and bottom of the grid. For the grid shown in Figure \ref{fig:2drep}(a), we randomly select one image that falls into each block, and display the image in its corresponding block. If there is no image in the block, the block remains black. From Figure \ref{fig:2drep}(a), we  observe that the axes are not only aligned with their assigned concepts, they also incorporate joint information. For example, a ``person in bed'' has high activation on both the ``person'' axis and ``bed'' axis.

We also include a 2D histogram of the number of images that fall into each block. As shown in Figure \ref{fig:2drep}(b), most images are distributed near the center (which is the origin) suggesting that the samples' feature vector has high probability to be nearly orthogonal to the concept axes we picked (meaning that they do not exhibit the two concepts), and consequently the latent features have near $0$ activation on the concept axes themselves.

\subsubsection{Trajectory of Concepts in Different Layers} \label{sec:trajectory}

Although our objective is the same when we apply the CW module to different layers in the same CNN, the latent space we get might be different. This is because different layers might be able to express different levels of semantic meaning. Because of this, it might be interesting to track how the representation of a single image will change as the CW module is applied to different layers of the CNN.

In order to better understand the latent representation, we plot a 2D slice of the latent space. Unlike in the 2d-representation space visualization (Figure \ref{fig:2drep}), here, a point in the plot is not specified by the activation values themselves but by their rankings. For example, the point $(0.7, 0.1)$ means the point is at the $70^{th}$ percentile for the first axis and the $10^{th}$ percentile in the second axis. We use the percentage instead of using the value, because as shown in the 2d-representation space visualization (Figure \ref{fig:2drep}), most points are near the center of the plot, so the rankings spread the values for plotting purposes.

\begin{figure*}[t]
    \centering
    \includegraphics[width=170mm]{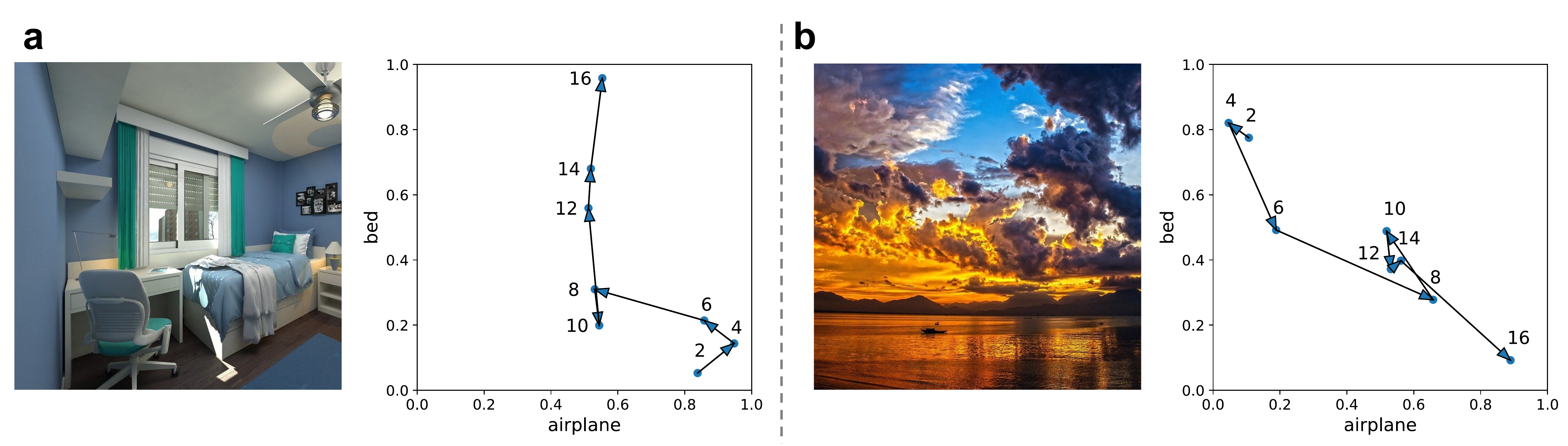}
    \caption{2D representation plot of two representative images. Each point in the right trajectory plot corresponds to the percentile rank for the activation values on each axis. The number labeling each point on the plot provides the layer depth of the CW module. The trajectory shows how the percentile rank of the left image changes when CW is applied to different layers. \label{fig:trajectory}}
\end{figure*}

Figure \ref{fig:trajectory} shows the 2D representation plot of two representative images. Each point in the plot corresponds to the percentile rank representation of the image when the CW module is applied to different layers. The points are connected by arrows according to the depth of the layer. These plots confirm that the abstract concepts learned in the lower layers tend to capture lower-level meaning (such as colors or shapes) while the higher layers capture high-level meaning (such as types of objects). For example, in the left image in Figure  \ref{fig:trajectory}(a), the bed is blue, where blue is typical low level information about the ``airplane'' class but not about the ``bed'' class since bedrooms are usually warm colors. Therefore, in lower layers, the bed image has higher ranking in the ``airplane'' axis than the ``bed'' axis. However, when CW is applied to deeper layers, high level information is available, and thus the image becomes highly ranked on the ``bed'' axis and lower on the ``airplane'' axis. 

In Figure \ref{fig:trajectory}(b), traversing through the networks' layers, the image of a sunset does not have the typical blue coloring of a sky. Its warm colors put it high on the ``bedroom'' concept for the second layer, and low on the ``airplane'' concept. However, as we look at higher layers, where the network can represent more sophisticated concepts, we see the image's rank grow on the ``airplane'' concept (perhaps the network uses the presence of skies to detect airplanes), and decrease on the ``bed'' concept.

\subsection{Separability of Latent Representations}
\label{sec:inter_inner}
In this subsection, we evaluate properties of the spatial distribution of the concepts in the latent space. By experimentally comparing such properties across latent representations produced by the CW module and other methods, we demonstrate that the issues arising in standard methods, as outlined in Section \ref{sec:whynotnn}, do not occur when using CW. We also investigate such properties on a non-posthoc neural network, trained with an auxiliary loss
that aims to classify different concepts in the latent space  (that is, in the objective, there are classification losses for each axis, using each axis' assigned concept as its label). Interestingly, we find that such issues mentioned in Section \ref{sec:whynotnn} may also exist in that network. The experiments in Section \ref{sec:inter_inner} were all done on ResNet18. The CW module was trained with seven simultaneous MS COCO concepts.

Specifically, for each concept image, we first extract its latent space representation. The representation for instance $j$ of concept $i$ is denoted $\mathbf{x}_{ij}$. Then, intra-concept similarity for concept $i$, denoted $d_{ii}$, is defined to be:
\begin{equation}
    d_{ii} = \frac{1}{n^2}\left(\sum_{j=1}^{n} \sum_{k=1}^{n} \frac{\x_{ij}\cdot \x_{ik}}{\|\x_{ij}\|_2\|\x_{ik}\|_2}\right)
\end{equation}
where $n$ is the total number of instances of concept $i$.

Inter-concept similarity between concept $p$ and $q$ is similarly defined as:
\begin{equation}
    d_{pq} = \frac{1}{nm}\left(\sum_{j=1}^{n} \sum_{k=1}^{m} \frac{\x_{pj}\cdot \x_{qk}}{{\|\x_{pj}\|_2\|\x_{qk}\|_2}}\right)
\end{equation}
where $n$ and $m$ are the number of instances of concepts $p$ and $q$ respectively. Indeed, intra-concept similarity is the average pairwise cosine similarity between instances of the same concept, and inter-concept similarity is the average pairwise cosine similarity between instances of two different concepts. 

\begin{figure*}[ht]
\centering
\includegraphics[width=170mm]{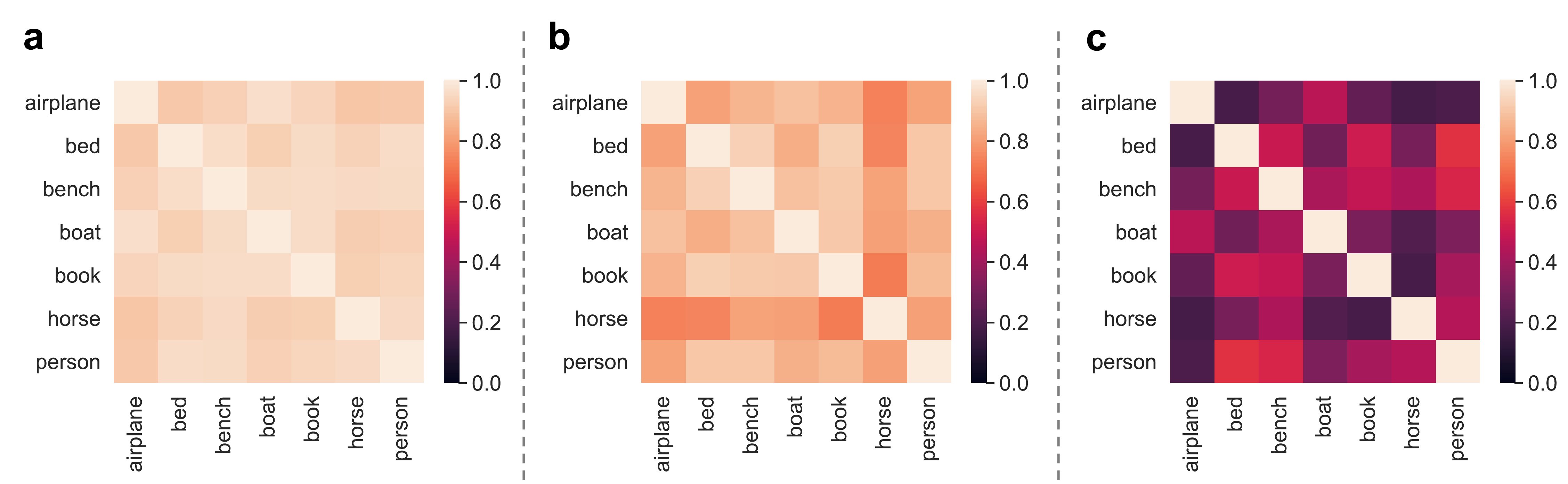}
\caption{Normalized intra-concept and inter-concept similarities. The diagonal values are normalized average similarities (see definition in Section \ref{sec:inter_inner}) between latent representations of images of the same concept; off-diagonal values are normalized average similarities between latent representations of images of different concepts. \textbf{a}, The $16^{th}$ layer is a BN module; \textbf{b}, The $16^{th}$ layer is a BN module with auxiliary loss to classify these concepts; \textbf{c}, The $16^{th}$ layer is a CW module. \label{fig:inner_product}}
\end{figure*}

With those defined, we plot heat maps in Figure \ref{fig:inner_product} where value in cell at row $i$ column $j$ is computed as:
\begin{equation}
    Q_{ij} = \frac{d_{ij}}{\sqrt{d_{ii}d_{jj}}}.
\end{equation}
From Figure \ref{fig:inner_product}, we notice that with the CW module, latent representations of concepts achieve greater separability: the ratios between inter-concept and intra-concept similarities (average 0.35) are notably smaller that of standard CNNs (average 0.94). In addition, without normalization, the CW module has very small inter-concept similarities (average 0.05) while analogous values for a standard neural network are around 0.74. This means that in the latent space of CW, two concepts are nearly orthogonal, while in a standard neural network, they are generally not. \textit{This indicates that some of the problems we identified in Section \ref{sec:whynotnn} occur in standard neural networks, but they do not occur with CW.}

In this experiment, as mentioned earlier, we also trained a standard neural network with a concept-distinction auxiliary loss. The auxiliary loss is the cross entropy of the first several dimensions in the latent space with respect to the concepts we investigated. Shown in Figure \ref{fig:inner_product}(b), the latent representations do not naturally help concept separation. The average ratio between inter-concept and intra-concept similarities is 0.85. Without normalization, the average inter-concept similarity is also around 0.74, similar to that of the standard neural network without the auxiliary loss. This has important implications: \textit{good discriminative power in the latent space does not guarantee orthogonality of different concepts. Thus, the whitening step is crucial for representing pure concepts.}

\subsection{Quantitative Evaluation of Interpretability}
\label{sec:quantitative_eval}

In this subsection, we measure the interpretability of the latent space quantitatively and compare it with other concept-based methods. 

First, we measure the purity of learned concepts by the AUC (of classifying the concept, not classifying with respect to the label for the overall prediction problem) calculated from the activation values. To calculate the test AUC, we divide the concept bank, containing 80 concepts extracted from MS COCO, into training sets and test sets. After training the CW module using the training set, we extract the testing samples' activation values on the axis representing the concept. For the target concept, we assign samples of this concept to the label $1$ while giving samples of the other 79 concepts label $0$. In this way, we calculate the one-vs-all test AUC score of classifying the target concept in the latent space. The AUC score measures whether the samples belonging to a concept are ranked higher than other samples. That is, the AUC score indicates the purity of the concept axis. Specifically, we randomly choose 14 concepts from the concept bank for the purity comparison. Since our CW module can learn multiple concepts at the same time, we divide the 14 concepts into two groups and train CW with 7 simultaneous concept datasets.

\begin{figure}[ht]
    \centering
    \includegraphics[width=80mm]{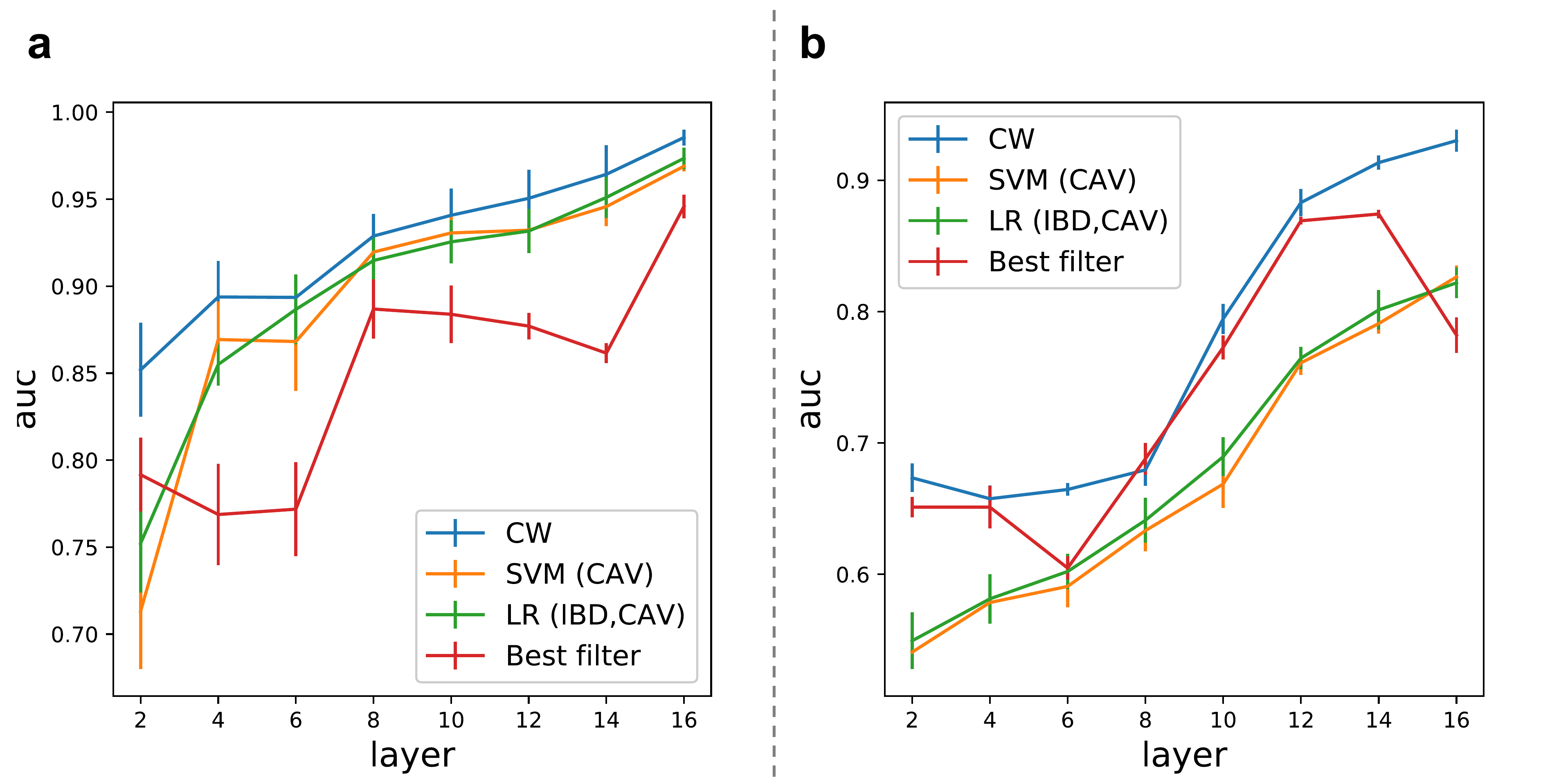}
    \caption{Concept purity measured by AUC score. \textbf{a}, concept ``airplane''; \textbf{b}, concept ``person.'' Concept purity of CW module is compared to several posthoc methods on different layers.  The error bar is the standard deviation over 5 different test sets, and each one is $20\%$ of the entire test set. \label{fig:auc}}
\end{figure}

\begin{figure}[ht]
    \centering
    \includegraphics[width=80mm]{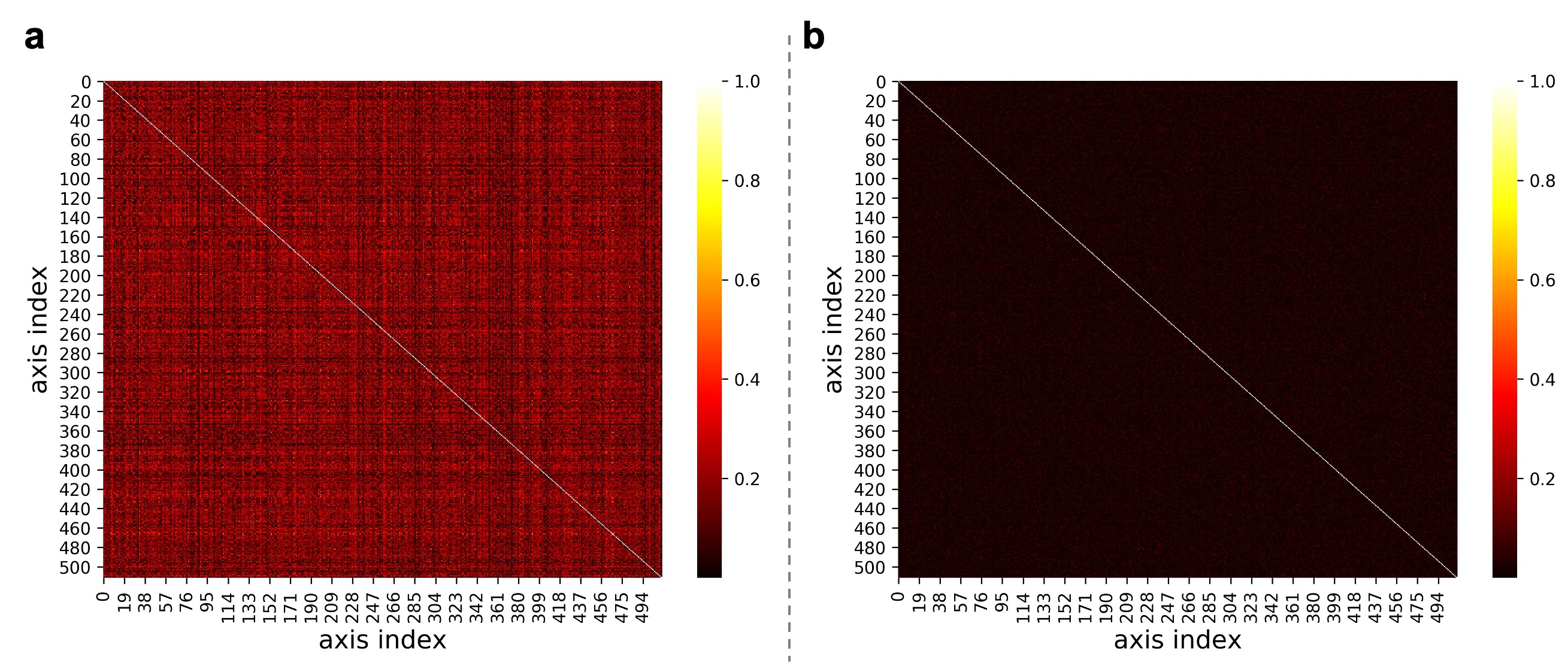}
    \caption{Absolute correlation coefficient of every feature pair in the $16^{th}$ layer.\textbf{a}, when the $16^{th}$ layer is a BN module; \textbf{b}, when $16^{th}$ layer is a CW module. \label{fig:correlation}}
\end{figure}

We compared the AUC concept purity of CW with the concept vectors learned by TCAV \cite{kim2018interpretability} from black box models, IBD \cite{zhou2018interpretable} from black box models, and filters in standard CNNs \cite{zhou2014object}. Since TCAV and IBD already find concept vectors, we use the samples' projections on the vectors to measure the AUC score. Note that in their original papers, the concept vectors are calculated for only one concept each time; therefore, we calculated 14 different concept vectors, each by training a linear classifier in the black box's latent space, with the training set of the target concept as positive samples and samples randomly drawn from the main dataset as negative samples. For standard CNNs, we measure the AUC score for the output of all filters and choose the best one to compare with our method, separately for each concept (denoted ``Best Filter''). Figure \ref{fig:auc} shows the AUC concept purity of ``airplane'' and ``person'' of these methods across different layers. The error bars on Figure \ref{fig:auc} were obtained by splitting the testing set into 5 parts and calculating AUC over each of them. The AUC plots for the other 12 concepts are shown in Supplementary Information \ref{sec:auc_all}. From the plots, we observe that \textit{concepts learned in the CW module are generally purer than those of other methods.} This is accredited to the orthogonality of concept representations as illustrated in Section \ref{sec:whynotnn}, as a result of CW's whitening of the latent space and optimization of the loss function.
We perform another quantitative evaluation that aims to measure the correlation of axes in the latent space before and after the CW module is applied. For comparison with posthoc methods like TCAV and IBD, we measure the output of their BN modules in the pretrained model, because the output of these layers are mean centered and normalized, which, as we discussed, are important properties for concept vectors. Shown by the absolute correlation coefficients plotted in Figure \ref{fig:correlation}(a), the axes still have relatively strong correlation after passing through the BN module. If CW were applied instead of BN, they would instead be decorrelated as shown in Figure \ref{fig:correlation}(b). This figure shows the correlation matrices for the $16^{th}$ layer. The same correlation comparison is shown in Supplementary Information \ref{sec:correlation_all} when CW is applied to other layers. These results reflect why purity of concepts is important; \textit{when the axes are pure, the signal of one concept can be concentrated only on its axis, while in standard CNNs, the concept could be strewn throughout the latent space.}

\subsection{Concept Importance}
\label{sec:concept_importance}

In order to obtain practical insights for how the concepts contribute to the classification results, we can measure the concept importance. The concept importance of the $j^{th}$ axis is defined as the ratio of a ``switched loss'' to the original loss:
\begin{equation}
    CI_j = \frac{e^{(j)}_{\rm switch}}{e_{\rm orig}}
\end{equation}
where the switched loss $e^{(j)}_{\rm switch}$ is the loss calculated when the sample values of $j^{th}$ axis are randomly permuted, and $e_{\rm orig}$ is the original loss without permutation. The expression for $CI_j$ is similar to classical definitions of variable importance  \cite{breiman2001random,fisher2019all}. Specifically:
\begin{itemize}
\item To measure the contribution of a concept to the entire classifier, the training loss function can be used in the variable importance calculation, which is the multi-class cross entropy in this case. \item To measure the contribution of a concept to a target class, e.g., how much ``bed'' contributes to ``bedroom,'' one can use a balanced binary cross entropy loss in the variable importance calculation, calculated on the softmax probability of the target class. The concept importance score is measured on the test set to prevent overfitting.
\end{itemize}

\begin{figure}[t]
    \centering
    \includegraphics[width=80mm]{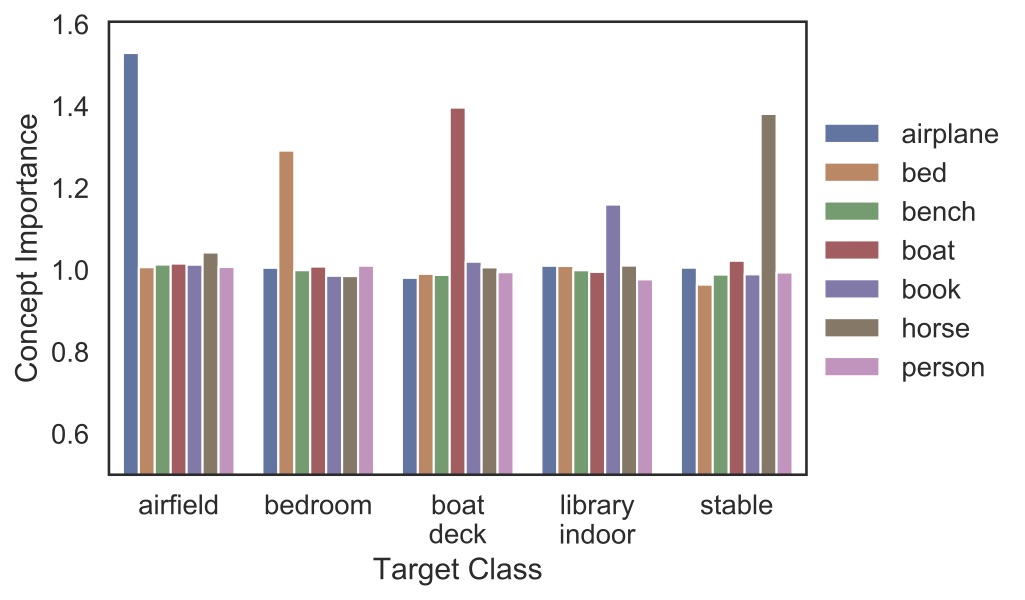}
    \caption{Concept importance to different Places365 classes measured on the concept axes when CW is applied to the $16^{th}$ layer. Each group in the bar plot corresponds to a target class. The bars in the same group show the concept importance scores of the learned concepts to the target class. Concepts that relate meaningfully to the target class (e.g., ``airplane'' and ``airfield'') have larger importance scores than irrelevant concepts.\label{fig:concept_importance}}
\end{figure}

In our experiments, we measure concept importance scores of the learned concepts to different target classes in the Places365 dataset (corresponding to the second of the bullets above). Figure \ref{fig:concept_importance} shows the results in a grouped bar plot. The target classes we choose relate meaningfully to a specific concept learned in CW (e.g., ``airplane'' and ``airfield''). We apply CW on the $16^{th}$ layer since the concepts are generally purer in the layer, as shown in Figure \ref{fig:auc}. As shown in Figure \ref{fig:concept_importance}, the irrelevant concepts have concept importance scores near 1.0 (no contribution), e.g., ``airplane'' is not important to the detection of ``bedroom.'' For the concepts that relate meaningfully to the target class, e.g., ``airplane'' to ``airfield,'' the concept importance scores are much larger than those for other concepts. \textit{Thus, the concept importance score measured on the CW latent space can tell us the contribution of the concept to the classification. For example, it can tell us how much a concept (such as ``airplane'' contributes to classifying  ``airfield,'' or how much ``book'' contributes to classifying ``library.''}

\subsection{Case Study: Skin Lesion Diagnosis}
\label{sec:isic}
We provide a case study of a medical imaging dataset of skin lesions. The dataset of dermoscopic images is collected from the ISIC archive \cite{isic2020}. Because the dermoscopic images corresponding to different diagnoses vary greatly in appearance, we focus on predicting whether a skin lesion is malignant for each of the histopathology images (9058 histopathology images in total). We choose ``age $<$ 20'' and ``size $\geq$ 10 mm'' as the concepts of interest and select the images with corresponding meta information to form the concept datasets. We chose these concepts due to their availability in the ISIC dataset. The cutoff, for instance, of 10mm is used commonly for evaluation of skin lesions \cite{lewis1998}. Details about the experimental results 
including test accuracy, separability of latent representation, AUC concept purity, correlation of axes, and concept importance are shown in Supplementary Information \ref{sec:isic_app}. The main results of the case study are: 
\begin{itemize}
\item The conclusions of CW performance analysis on the ISIC dataset are very similar to our earlier conclusions on the Places dataset, in terms of main objective test accuracy, separability of latent representation, AUC concept purity, and correlation of axes. \item Concept importance scores measured on the CW latent space can provide practical insights on which concepts are potentially more important in skin lesion diagnosis.
\end{itemize}

\section{Conclusion and Future Work}
\label{sec:conclusion}
Concept whitening is a module placed at the bottleneck of a CNN, to force the latent space to be disentangled, and to align the axes of the latent space with predefined concepts. By building an inherently interpretable CNN with concept whitening, we can gain intuition about how the network gradually learns the target concepts (or whether it needs them at all) over the layers without harming the main objective's performance.

There are many avenues for possible future work. Since CW modules are useful for helping humans to define primitive abstract concepts, such as those we have seen the network use at early layers, it would be interesting to automatically detect and quantify these new concepts (see ref. \cite{ghorbani2019towards}). Also the requirement of CW to completely decorrelate the outputs of all the filters might be too strong for some tasks. This is because concepts might be highly correlated in practice such as ``airplane'' and ``sky'' In this case, we may want to soften our definition of CW. We could define several general topics that are uncorrelated, and use multiple correlated filters to represent concepts within each general topic. In this scenario, instead of forcing the gram matrix to be the identity matrix, we could make it block diagonal. The orthogonal basis would become a set of orthogonal subspaces.

\section*{Data Availability}
All datasets that support the findings are publicly available, including Places365 at \href{http://places2.csail.mit.edu}{http://places2.csail.mit.edu}, MS COCO at \href{https://cocodataset.org/}{https://cocodataset.org/} and ISIC at \href{https://www.isic-archive.com}{https://www.isic-archive.com}.

\section*{Code Availability}
The code for replicating our experiments is available on \href{https://github.com/zhiCHEN96/ConceptWhitening}{https://github.com/zhiCHEN96/ConceptWhitening} (\href{https://doi.org/10.5281/zenodo.4052692}{https://doi.org/10.5281/zenodo.4052692}).

\bibliography{ref}

\renewcommand\refname{Supplementary References}
\renewcommand{\figurename}{Supplementary Figure}
\renewcommand\tablename{Supplementary Table}

\newpage

\appendix
\onecolumn
\title{Concept Whitening for Interpretable Image Recognition \\ Supplementary Information}
\date{}
\maketitle
\section{Concept Activation Calculation and Concept Activation Comparison Experiments}
\label{sec:activation}
\subsection{Calculations of Concept Activation Based on Feature Maps}
\label{sec:activation_defs}
The output of a single filter is a $h\times w$ feature map. However, a scalar is needed to quantify how much a sample is activated on a concept, which is used in both optimization and evaluation. Based on a feature map, multiple reasonable ways exists to calculate the concept activation.

\begin{figure}[htbp]
    \includegraphics[width=6.75in]{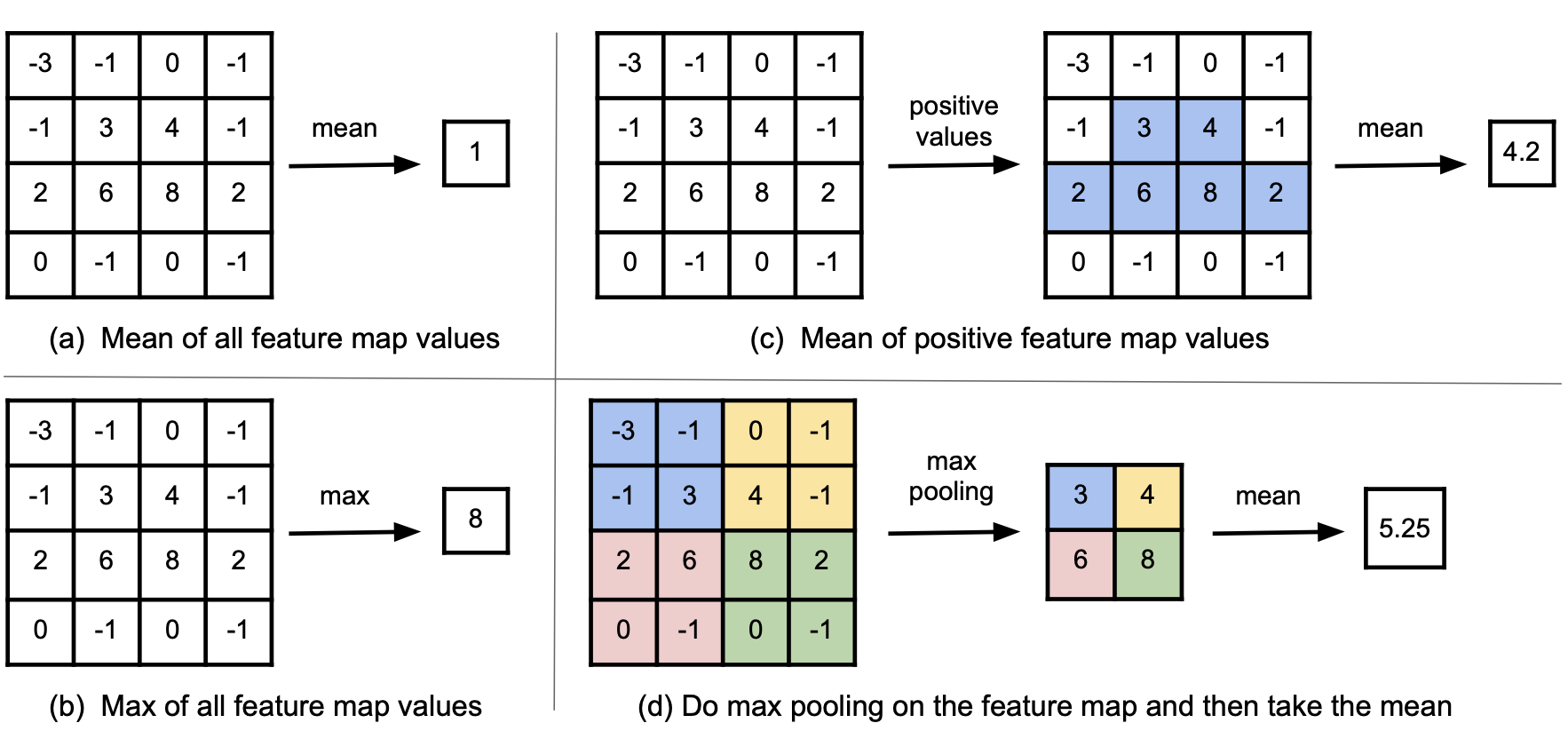}
  \caption{Four methods of calculating concept activation based on the feature map.}
  \label{fig:activation_defs}
\end{figure}

Specifically, we try the following calculations to produce an activation value:
\begin{itemize}[topsep=0pt,noitemsep]
    \item Mean of all feature map values
    \item Max of all feature map values
    \item Mean of all positive feature map values
    \item Mean of down-sampled feature map obtained by max pooling.
\end{itemize}
Supplementary Figure  \ref{fig:activation_defs} shows these four methods of calculating the activation through demonstration. Among them, the mean of values is more suitable for capturing low-level concepts since they are distributed throughout the feature map. For high-level concepts, the max value and mean of positive values are more powerful: they can capture high-level concepts such as objects, since objects usually occur just in one location, not repeatedly throughout an image. The mean of max-pooled values is a combination of the previous types and is capable of representing both high-level and low-level concepts. Intuitively, the mean of max pooled values is more similar to the max function when applied to higher layers and more similar to the mean function when applied to lower layers. This is because, for higher layers, the mean is taken of only a few values, simply because higher layers are smaller in size. Thus, the max is the dominant calculation. In contrast, for lower layers, which are much larger, the max's are taken over a relatively small number of elements (local regions), and then the mean is taken over all of the local regions. Hence the mean is the dominant calculation for lower layers. 

\subsection{Top-10 Activated Images Based on Different Calculations}
\label{sec:top10_act}
\begin{figure}[t]
    \subfigure[Mean of all feature map values]{
    \centering
    \begin{minipage}[t]{1\linewidth}
    \includegraphics[width=6.75in]{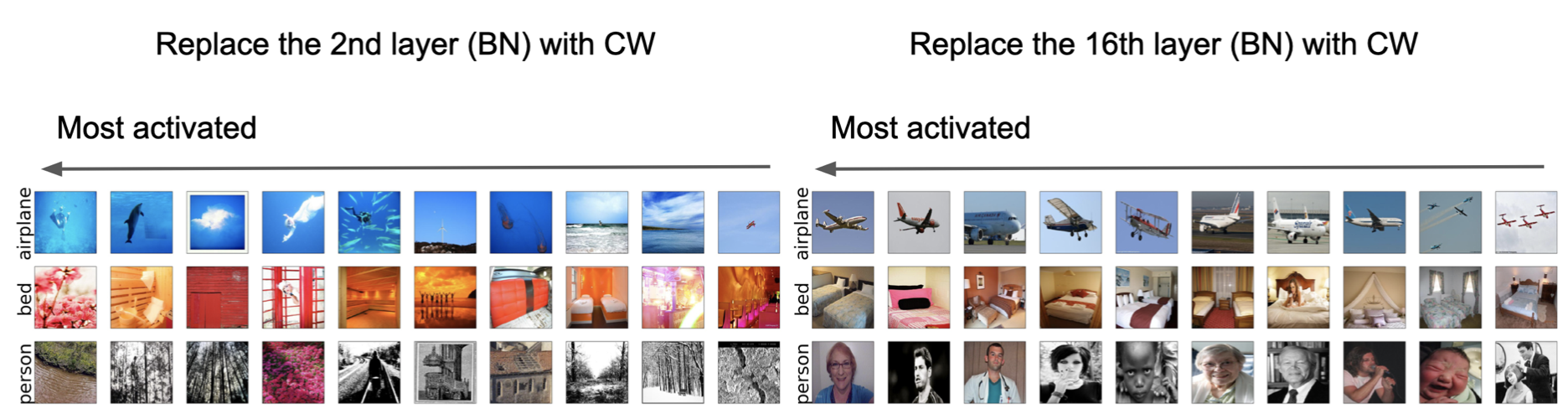}
    \end{minipage}%
    } 
    \subfigure[Max of all feature map values]{
    \centering
    \begin{minipage}[t]{1\linewidth}
    \includegraphics[width=6.75in]{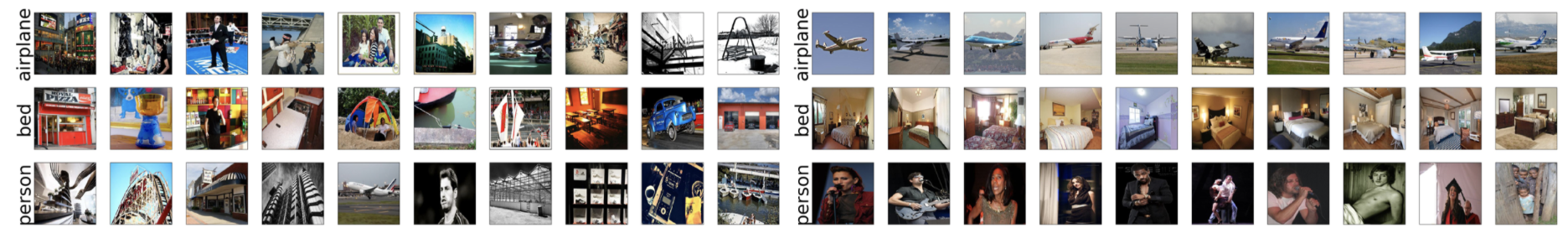}
    \end{minipage}%
    } 
    \subfigure[Mean of all positive feature map values]{
    \centering
    \begin{minipage}[t]{1\linewidth}
    \includegraphics[width=6.75in]{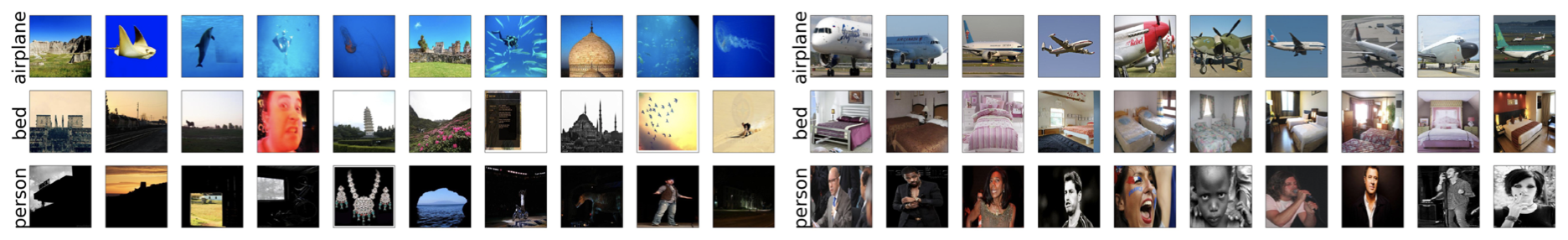}
    \end{minipage}%
    } 
    \subfigure[Mean of down-sampled feature map obtained by max pooling]{
    \centering
    \begin{minipage}[t]{1\linewidth}
    \includegraphics[width=6.75in]{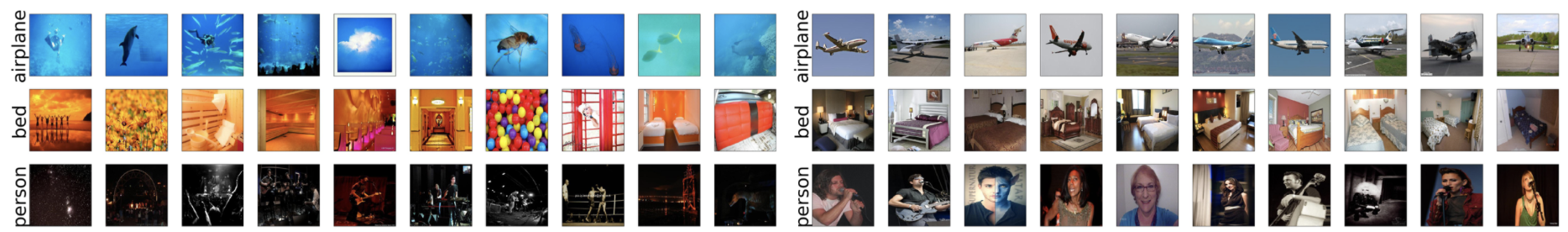}
    \end{minipage}%
    } 
  \caption{Top-10 activated images obtained by different calculations of concept activation. Depending on which choice (max, mean, mean of positives, mean of max pool values), different abstract concepts are generated in the second layer. For instance, for the person class, using the mean calculation on the second layer (top left), the abstract concept is a dense texture. For the person class using the mean of max-pooled values (bottom left), the abstract concept is a dark background with vertical lights. For the bed concept with the mean of max-pooled values (bottom left), the abstract concept is warm colors, whereas for the max calculation (second row left) the abstract concept seems to be related to boundaries of different colors. These concepts could later be formalized, if desired, to create better or more interpretable classifiers in the future.}
  \label{fig:top10_act}
\end{figure}

Supplementary Figure  \ref{fig:top10_act} shows the top-10 activated images under the four different calculations for concept activation. The CNN architecture, dataset and the depth of the CW module are the same as before. The figures show that when concept activation is calculated in different ways, the most activated images may look different and the network even may discover completely different lower-level characteristics. For example, when CW is applied to the $2^{nd}$ layer, the network discovered the lower-level characteristics of the concept ``bed'' to be warm colors when the activation was the mean of feature map values, while the lower-level characteristics seems to involve boundaries of colors if activation is calculated as the max value. Also if the activation is calculated as the mean of all values, the ``person'' concept gives rise to dense texture, while under the mean of max-pooled values, the ``person'' concept is characterized as a dark background with vertical lights. This difference in the discovered lower-level characteristics could be explained by the fact that these calculation methods focus on different locations within the image: the mean value focuses on the whole image while the max value only looks at one place within the image.
\subsection{Concept AUC Based on Different Activation Calculations}
\label{sec:auc_act}
Supplementary Table  \ref{fig:auc_act} shows concept AUC when different concept activation definitions are used. The definition and calculation of concept AUC is the same as in the main paper. The dataset and CNN architecture are also the same. To compare these concept activations' capability to capture both high-level concepts and low-level concepts, we apply CW to the $2^{nd}$ and $16^{th}$ layers of ResNet18. Supplementary Table  \ref{fig:auc_act} indicates that in the $2^{nd}$ layer, the max value of the feature map performs worse on AUC 
than the other calculation methods for two out of our three concepts. In contrast, 
in the $16^{th}$ layer, the mean performs poorly compared to the other methods. The max-pool-mean method performs well on both layers, for all concepts. This result matches our intuitive reasoning that the max-pool-mean combines the advantages of mean and max. It is suitable for capturing both low-level concepts and high-level concepts.
\begin{table}[htbp]
    \centering
    \begin{tabular}{lcc c cc c cc}
     \hline
     & \multicolumn{2}{c}{AUC-``airplane''} && \multicolumn{2}{c}{AUC-``bed''} && \multicolumn{2}{c}{AUC-``person''} \\
     \cline{2-3} \cline{5-6} \cline{8-9}
      & $2^{nd}$ layer & $16^{th}$ layer && $2^{nd}$ layer & $16^{th}$ layer && $2^{nd}$ layer & $16^{th}$ layer\\ 
     \hline
     Mean & 0.820 & 0.981 && 0.687 & 0.853 && 0.714 & 0.918\\ 
     Max & 0.716 & 0.992 && 0.589 & 0.904 && 0.759 & 0.969\\ 
     Positive-mean & 0.798 & 0.992 && 0.614 & 0.924 && 0.757 & 0.968\\ 
     Max-pool-mean & 0.818 & 0.993 && 0.692 & 0.906 && 0.757 & 0.966\\ 
     \hline
    \end{tabular}
    \caption{Concept AUC obtained by different calculations of concept activation. Max-pool-mean performs well when CW is applied both to low and high layers. }
    \label{fig:auc_act}
\end{table}
\section{Sensitivity Analysis of Main Objective Accuracy}
\label{sec:acc_sensitivity}

\subsection{Main Objective Accuracy when CW is Applied to Different Layers}
\label{sec:acc_all}
As mentioned in the main paper, we measure the main objective accuracy when CW applied to different layers. Tables \ref{fig:acc_vgg16} through  \ref{fig:acc_resnet18} show the layer-wise test accuracy of different CNN architectures. The dataset and CNN architectures are the same as in the main paper. Results in Tables \ref{fig:acc_vgg16} through \ref{fig:acc_resnet18} indicate that no matter which layer we apply CW, accuracy is not substantially impacted.
\begin{table*}[t]
\begin{floatrow}
\capbtabbox{
 \begin{tabular}{ccc}
 \hline
 CW layer & Top-1 acc. & Top-5 acc. \\
 \hline
 $1^{nd}$ & 53.2 & 83.8 \\
 $2^{th}$ & 53.3 & 83.8 \\
 $3^{th}$ & 53.4 & 83.8 \\
 $4^{th}$ & 53.4 & 83.9 \\
 $5^{th}$ & 53.2 & 83.9 \\
 $6^{th}$ & 53.3 & 83.8 \\
 $7^{th}$ & 53.5 & 83.8 \\
 $8^{rd}$ & 53.3 & 83.9 \\
 $9^{nd}$ & 53.4 & 83.8 \\
 $10^{th}$ & 53.2 & 83.8 \\
 $11^{nd}$ & 53.2 & 83.9 \\
 $12^{th}$ & 53.3 & 83.7 \\
 \hline
 \end{tabular}
}{
 \caption{Top-1 and top-5 accuracy of VGG16-CW on Places365 dataset. Our results indicate that the choice of layer to apply CW does not have a practical impact on accuracy.}
 \label{fig:acc_vgg16}
}

\capbtabbox{
 \begin{tabular}{ccc}
 \hline
 CW layer & Top-1 acc. & Top-5 acc. \\
 \hline
 $2^{nd}$ & 55.2 & 85.4 \\
 $5^{th}$ & 55.3 & 85.5 \\
 $8^{th}$ & 55.3 & 85.5 \\
 $11^{th}$ & 55.2 & 85.5 \\
 $14^{th}$ & 55.3 & 85.5 \\
 $17^{th}$ & 54.8 & 85.2 \\
 $20^{th}$ & 54.7 & 85.0 \\
 $23^{rd}$ & 54.8 & 85.0 \\
 $26^{nd}$ & 54.7 & 85.0 \\
 $29^{th}$ & 54.8 & 85.0 \\
 $32^{nd}$ & 54.8 & 85.1 \\
 $35^{th}$ & 54.7 & 85.0 \\
 $38^{th}$ & 54.8 & 85.1 \\
 $41^{st}$ & 54.6 & 85.0 \\
 $44^{th}$ & 54.7 & 84.9 \\
 $47^{th}$ & 54.6 & 85.0 \\
 \hline
 \end{tabular}
}{
 \caption{Top-1 and top-5 accuracy of ResNet50-CW on Places365 dataset. Our results indicate that the choice of layer to apply CW does not have a practical impact on accuracy.}
 \label{fig:acc_resnet50}
}

\end{floatrow}
\begin{floatrow}

\capbtabbox{
 \begin{tabular}{ccc}
 \hline
 CW layer & Top-1 acc. & Top-5 acc. \\
 \hline
 $14^{th}$ & 55.6 & 85.7 \\
 $39^{th}$ & 55.5 & 85.5 \\
 $88^{nd}$ & 55.5 & 85.6 \\
 $161^{th}$ & 55.5 & 85.6 \\
 \hline
 \end{tabular}
}{
 \caption{Top-1 and top-5 accuracy of DenseNet161-CW on Places365 dataset. Our results indicate that the choice of layer to apply CW does not have a practical impact on accuracy.}
 \label{fig:acc_densenet161}
}

\capbtabbox{
 \begin{tabular}{ccc}
 \hline
 CW layer & Top-1 acc. & Top-5 acc. \\
 \hline
 $2^{nd}$ & 53.9 & 84.2 \\
 $4^{th}$ & 54.0 & 84.5 \\
 $6^{th}$ & 54.0 & 84.3 \\
 $8^{th}$ & 54.0 & 84.2 \\
 $10^{th}$ & 54.0 & 84.3 \\
 $12^{th}$ & 53.9 & 84.1 \\
 $14^{th}$ & 53.7 & 83.9 \\
 $16^{th}$ & 53.5 & 83.8 \\
 \hline
 \end{tabular}
}{
 \caption{Top-1 and top-5 accuracy of ResNet18-CW on Places365 dataset. Our results indicate that the choice of layer to apply CW does not have a practical impact on accuracy.}
 \label{fig:acc_resnet18}
}

\end{floatrow}

\end{table*}

\subsection{Main Objective Accuracy versus Number of Concepts Trained in CW}
\label{sec:acc_ncpt}
We measure the main objective accuracy on Places365 when different numbers of concepts are trained within the CW module. The CNN architecture we evaluate is ResNet18. For each number of concepts, we average the result over three groups of randomly selected concepts. Also, for each group of simultaneous concepts, the result is averaged over different layers that CW is applied to. As shown in Supplementary Figure  \ref{fig:acc_ncpt}, both top-1 (Supplementary Figure  \ref{fig:acc_ncpt}(a)) and top-5 (Supplementary Figure  \ref{fig:acc_ncpt}(b)) accuracy are not significantly affected by the number of concepts. The drop of accuracy is less than $0.5\%$ when the number of concepts increases from 3 to 9.

\begin{figure}[ht]
    \subfigure[Top-1 test accuracy]{
    \centering
    \begin{minipage}[t]{0.45\linewidth}
    \centering
    \includegraphics[width=1.9in]{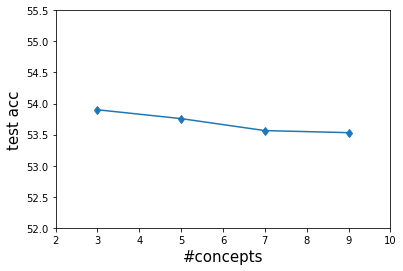}
    \end{minipage}%
    }
    \subfigure[Top-5 test accuracy]{
    \centering
    \begin{minipage}[t]{0.45\linewidth}
    \centering
    \includegraphics[width=1.9in]{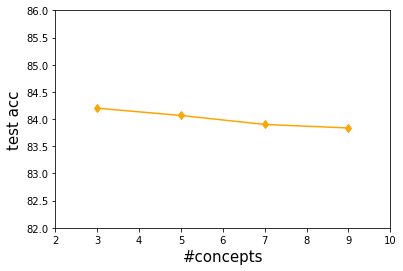}
    \end{minipage}%
    }
  \caption{Test accuracy on Places365 when different number of concepts are learned in CW. (a) Top-1 accuracy; (b) Top-5 accuracy.}
  \label{fig:acc_ncpt}
\end{figure}

\section{Correlation Matrix when CW is Applied to Different Layers}
\label{sec:correlation_all}
As is shown in the experiments, we calculate the correlations of axes in the latent space to quantitatively compare CW and other concept-based methods. Here, in Supplementary Figure  \ref{fig:correlation_all}, we present the absolute correlation coefficient matrices as heatmaps when these methods are applied to different layers ($2^{nd}$, $4^{th}$, $6^{th}$, $8^{th}$, $12^{th}$, $14^{th}$ and $16^{th}$ layer) in ResNet-18. The correlation coefficient is calculated on the test set. The darker the off-diagonal elements are, the more decorrelated the latent space is. Heatmaps of CW in Supplementary Figure  \ref{fig:correlation_all} are all near pure black. \textit{This demonstrates that CW can consistently decorrelate the latent space -- whichever layer it is applied on -- while the neural networks trained without any constraints can have strong correlations between different axes. Such a strongly decorrelated latent space enables the signal of one concept to be concentrated on one axis rather than throughout the latent space.} 
\begin{figure}[htbp]
    \centering
    \begin{minipage}[b]{1.0\linewidth}
        \centering
        \includegraphics[scale=0.36]{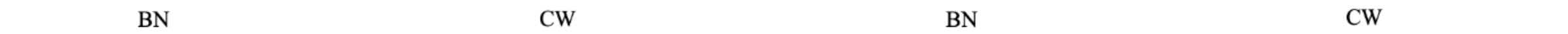}
    \end{minipage}
    \subfigure[$2^{nd}$ layer. left: BN, right: CW]{
    \begin{minipage}[b]{0.45\linewidth}
        \centering
        \scalebox{1.0}{
            \includegraphics[scale=0.24]{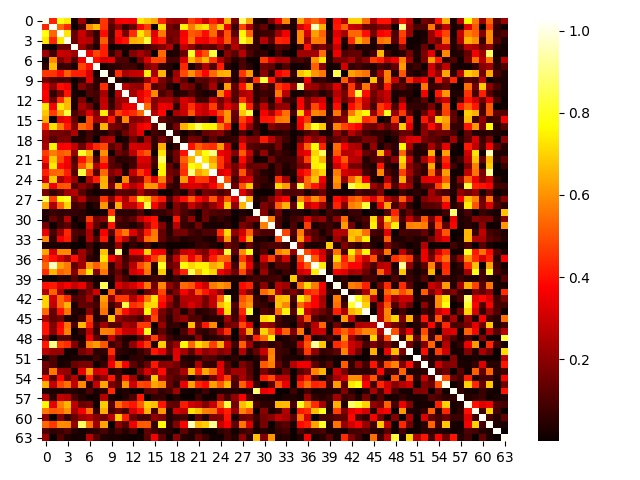}
            \includegraphics[scale=0.24]{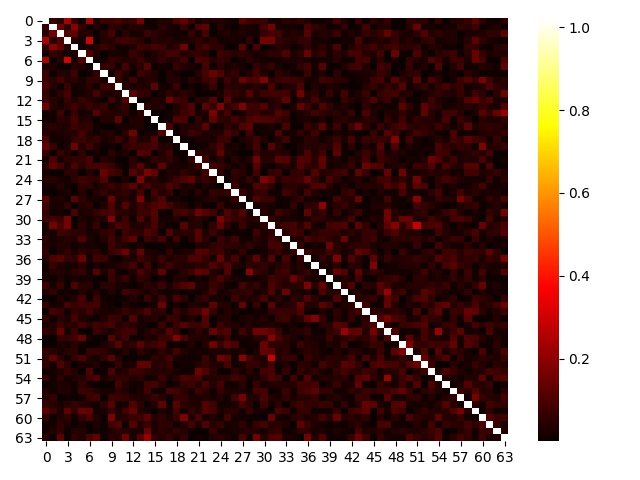}
        }
    \end{minipage}}
    \subfigure[$4^{th}$ layer. left: BN, right: CW]{
    \begin{minipage}[b]{0.45\linewidth}
        \centering
        \scalebox{1.0}{
            \includegraphics[scale=0.24]{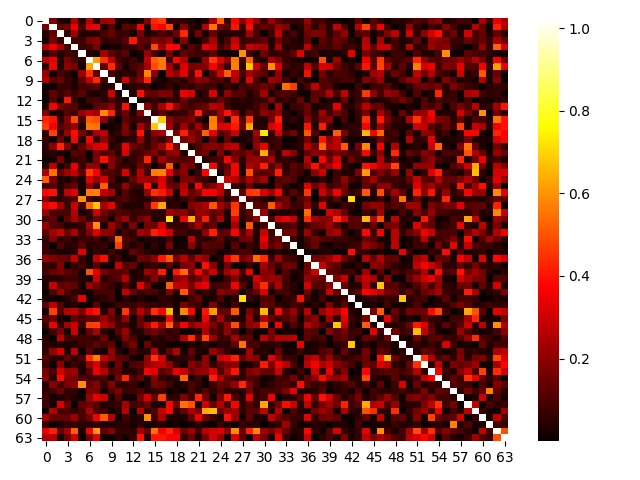}
            \includegraphics[scale=0.24]{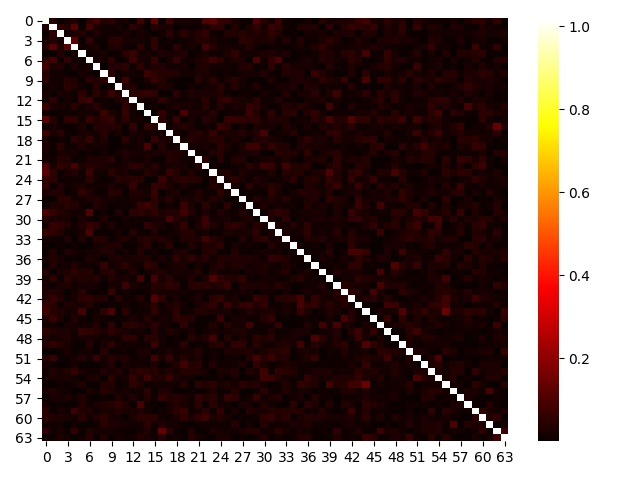}
        }
    \end{minipage}}
    \subfigure[$6^{th}$ layer. left: BN, right: CW]{
    \begin{minipage}[b]{0.45\linewidth}
        \centering
        \scalebox{1.0}{
            \includegraphics[scale=0.24]{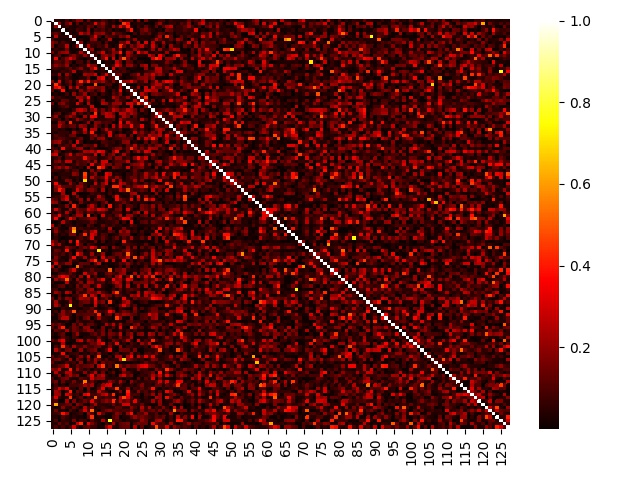}
            \includegraphics[scale=0.24]{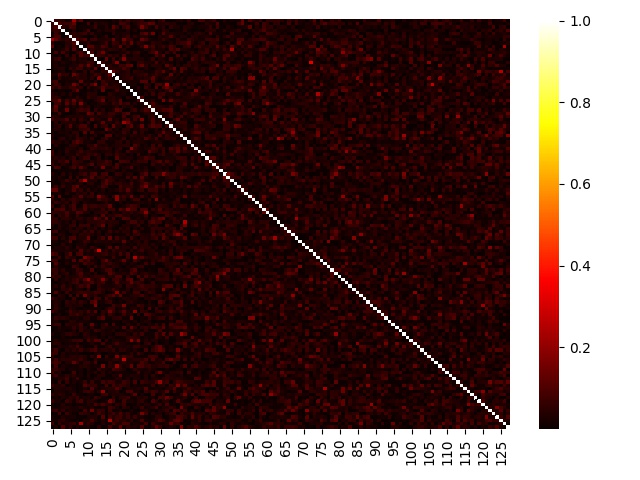}
        }
    \end{minipage}}
    \subfigure[$8^{th}$ layer. left: BN, right: CW]{
    \begin{minipage}[b]{0.45\linewidth}
        \centering
        \scalebox{1.0}{
            \includegraphics[scale=0.24]{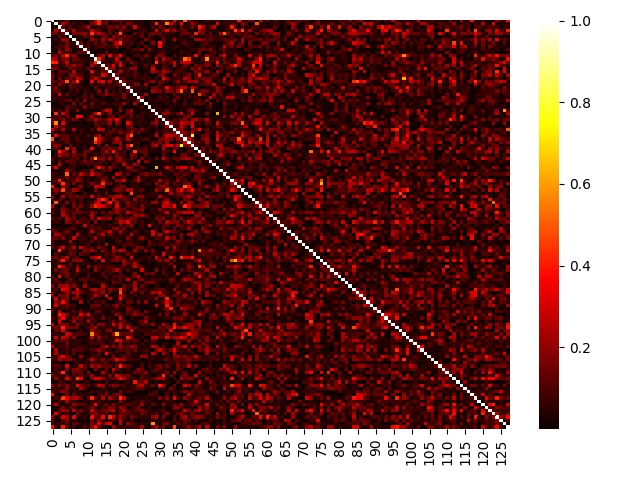}
            \includegraphics[scale=0.24]{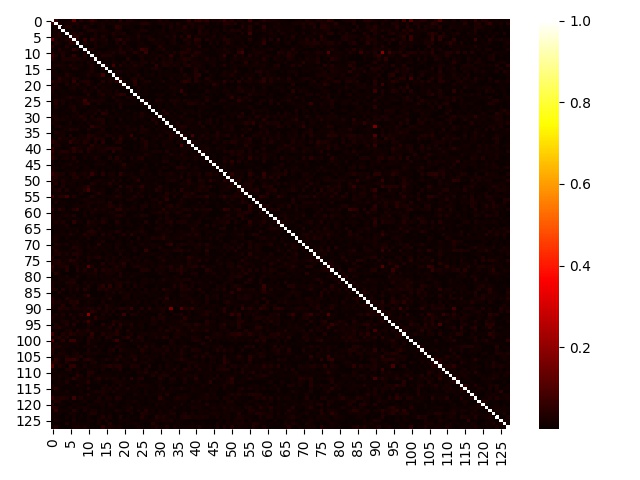}
        }
    \end{minipage}}
    \subfigure[$10^{th}$ layer. left: BN, right: CW]{
    \begin{minipage}[b]{0.45\linewidth}
        \centering
        \scalebox{1.0}{
            \includegraphics[scale=0.24]{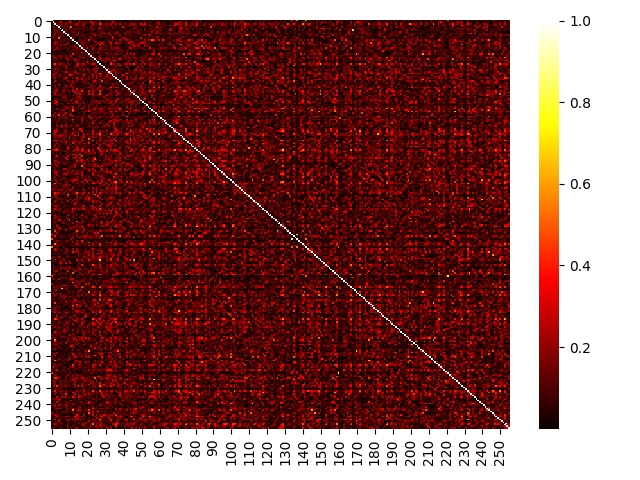}
            \includegraphics[scale=0.24]{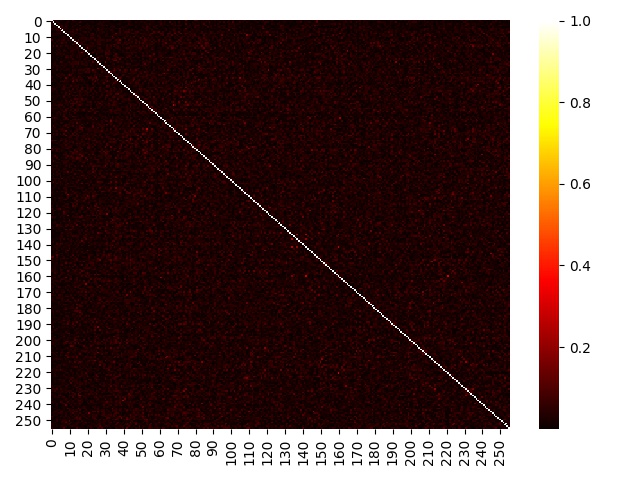}
        }
    \end{minipage}}
    \subfigure[$12^{th}$ layer. left: BN, right: CW]{
    \begin{minipage}[b]{0.45\linewidth}
        \centering
        \scalebox{1.0}{
            \includegraphics[scale=0.24]{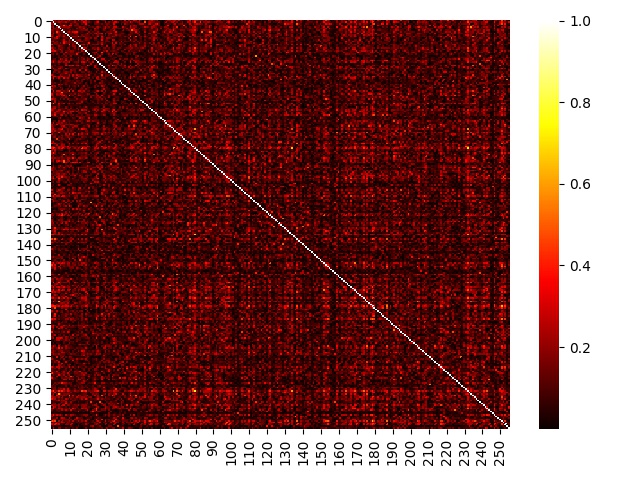}
            \includegraphics[scale=0.24]{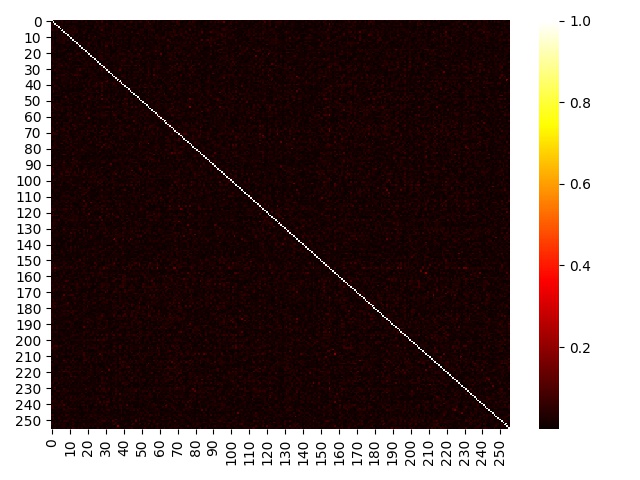}
        }
    \end{minipage}}
    \subfigure[$14^{th}$ layer. left: BN, right: CW]{
    \begin{minipage}[b]{0.45\linewidth}
        \centering
        \scalebox{1.0}{
            \includegraphics[scale=0.24]{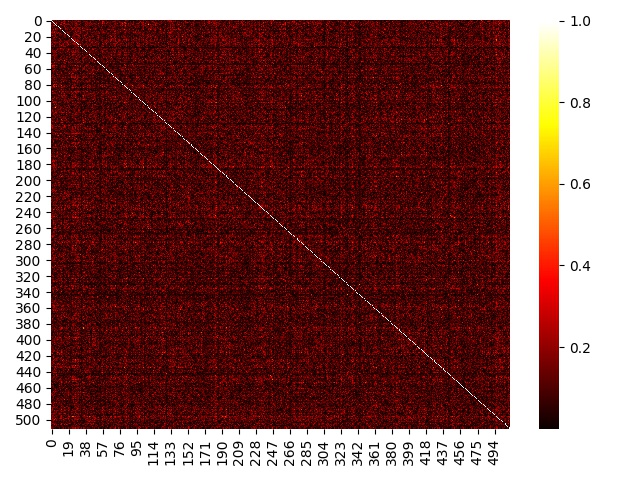}
            \includegraphics[scale=0.24]{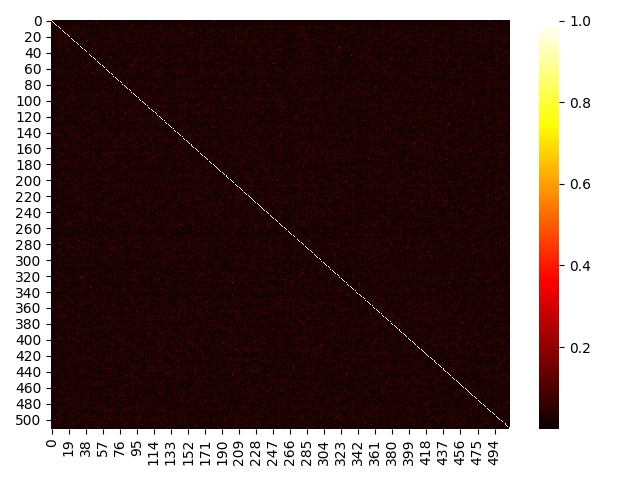}
        }
    \end{minipage}}
    \subfigure[$16^{th}$ layer. left: BN, right: CW]{
    \begin{minipage}[b]{0.45\linewidth}
        \centering
        \scalebox{1.0}{
            \includegraphics[scale=0.24]{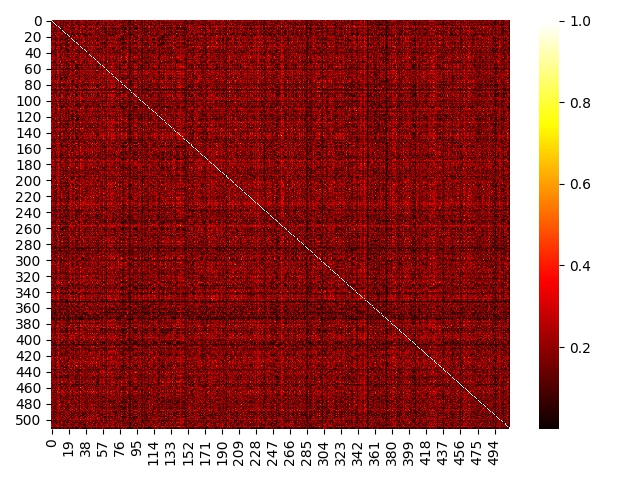}
            \includegraphics[scale=0.24]{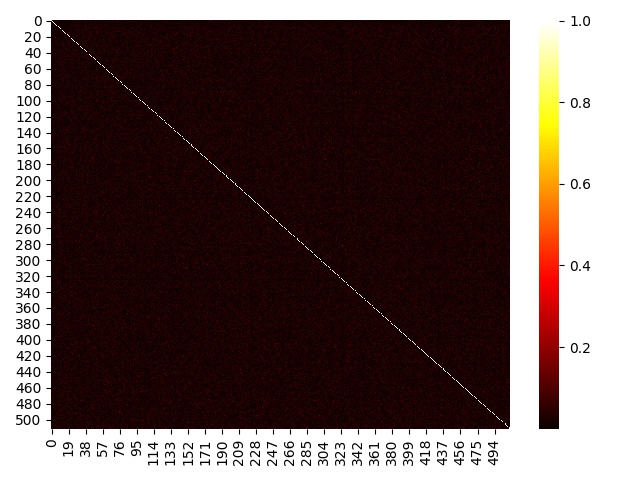}
        }
    \end{minipage}}
    \caption{
    Absolute correlation coefficient of every latent feature pair in the $2^{nd}$, $4^{th}$, $6^{th}$, $8^{th}$, $12^{th}$, $14^{th}$ and $16^{th}$ layer, calculated on the test set. For each pair of figures, the left figure is when the layer is a BN module; the right figure is when the layer is a CW module. For the CW module, the first several features represent the concepts. The correlations of CW are much lower off the diagonal, as desired. \label{fig:correlation_all}
}
\end{figure}

\section{Results on More Concepts}
\label{sec:more_concepts}
To show the capability of dealing with many concepts as well as the usefulness of the proposed method, we conduct experiments on more concepts, including concepts defined as objects (\ref{sec:more_object_concepts}) and concepts defined as general characteristics of objects and scenes (\ref{sec:more_general_concepts}).

\subsection{Object Concepts}
\label{sec:more_object_concepts}
 The object concept bank contains 80 concepts obtained from MS COCO \cite{lin2014microsoft} by cropping out objects in bounding boxes. We chose 7 concepts randomly selected from the concept bank each time, where the CW module was trained on all of these concepts at the same time.

\subsubsection{Top-10 Activated Images}
\label{sec:top10_all}
\begin{figure}[htbp]
    \subfigure[Concepts trained together: airplane, bed, bench, boat, book, horse, person]{
    \centering
    \begin{minipage}[t]{1\linewidth}
    \includegraphics[width=6.75in]{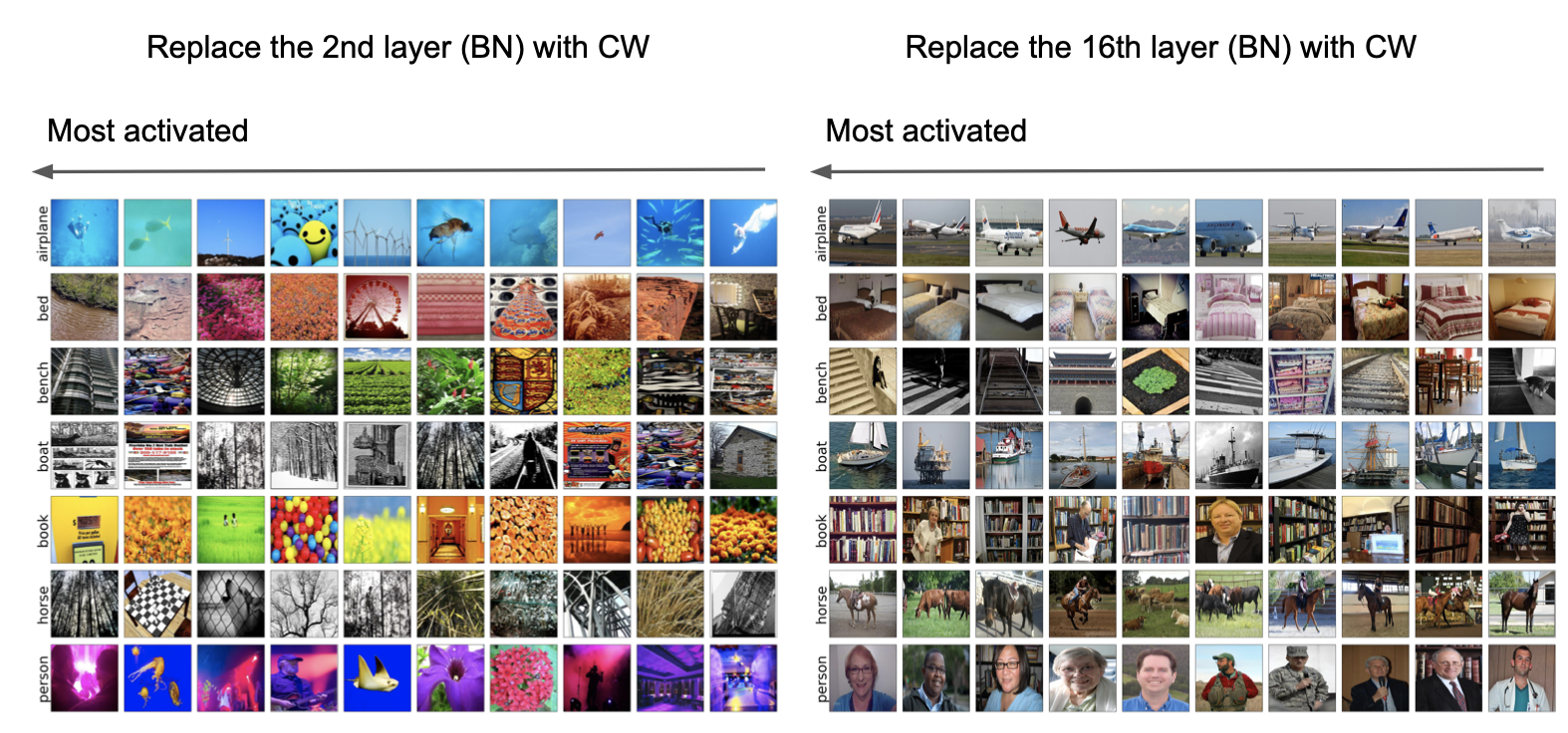}
    \end{minipage}%
    } 
    \subfigure[Concepts trained together: bus, car, dining table, potted plant, sink, umbrella, wine glass]{
    \centering
    \begin{minipage}[t]{1\linewidth}
    \includegraphics[width=6.75in]{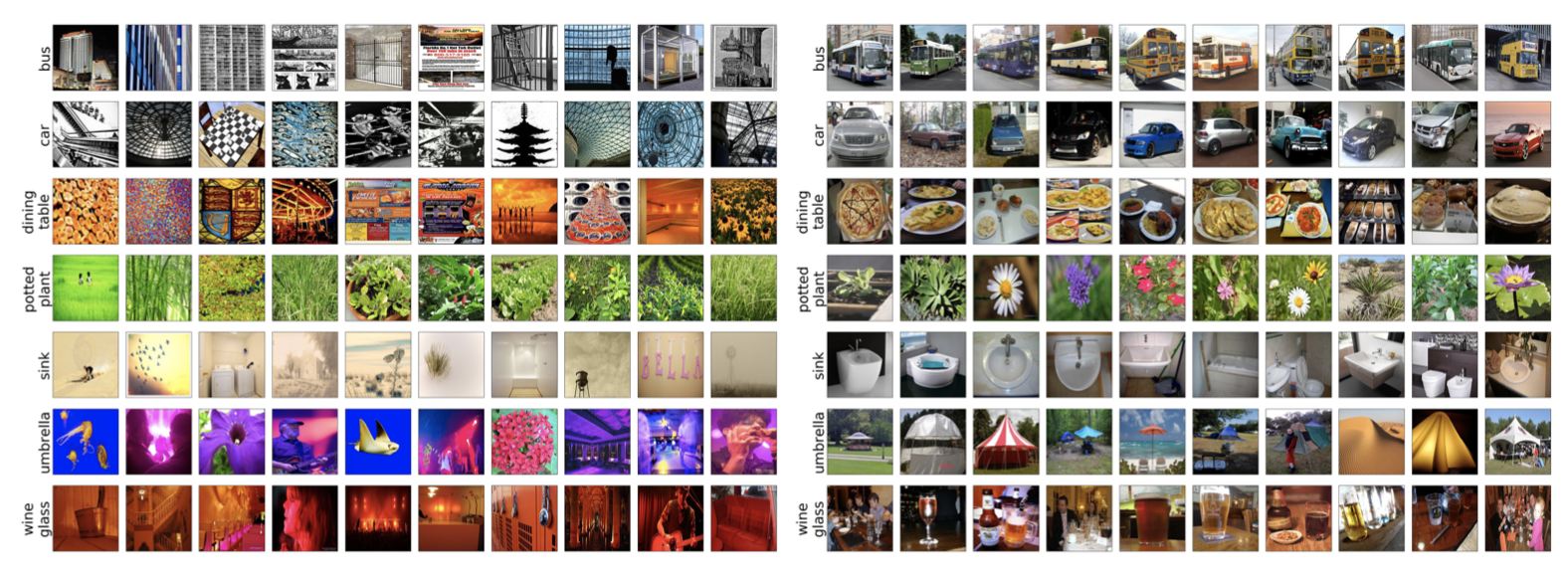}
    \end{minipage}%
    } 
  \caption{Top-10 activated images when CW is trained on more object concepts.}
  \label{fig:top10_all}
\end{figure}

Supplementary Figure  \ref{fig:top10_all} shows the two groups of top-10 activated images along the seven different concepts' axes. 
On the left of Supplementary Figure  \ref{fig:top10_all}, we can see that when the CW module is applied to a lower layer (the $2^{nd}$ layer), it captures some low-level information such as color and texture about the concept. Top activated images on the right of Supplementary Figure  \ref{fig:top10_all} demonstrate the concepts' high-level meaning when CW is located at a higher layer (the $16^{th}$ layer). 
An interesting finding is that when two similar concepts are given, for example ``bus'' and ``car,'' the network can learn their difference and distinguish them successfully in both low and high layers. Moreover, if images with the concept do not exist in the main dataset, the concept axes can be activated by images that are very similar to the concept, like slats and wood textures for the ``bench'' concept and tents for the ``umbrella'' concept. 

\subsubsection{AUC Concept Purity}
\label{sec:auc_all}
\begin{figure}[htbp] 
\centering
\subfigure[Concept ``airplane'']{
\begin{minipage}[t]{0.24\linewidth}
\centering
\includegraphics[width=1.5in]{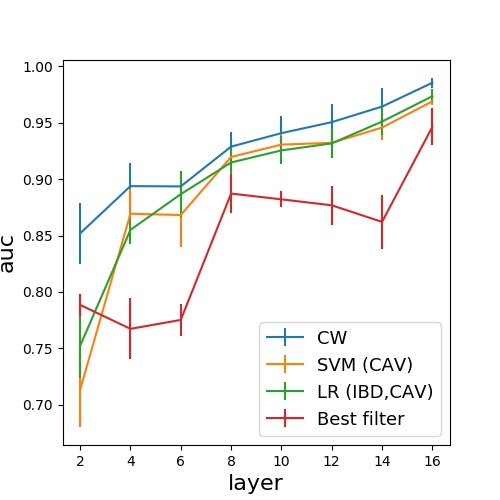}
\end{minipage}%
}
\subfigure[Concept ``bed'']{
\begin{minipage}[t]{0.24\linewidth}
\centering
\includegraphics[width=1.5in]{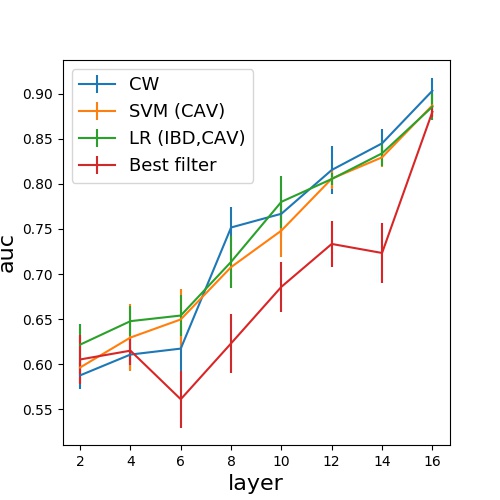}
\end{minipage}%
}
\subfigure[Concept ``bench'']{
\begin{minipage}[t]{0.24\linewidth}
\centering
\includegraphics[width=1.5in]{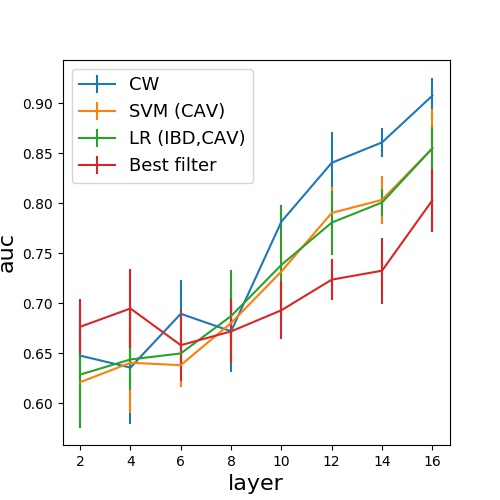}
\end{minipage}%
}
\subfigure[Concept ``boat'']{
\begin{minipage}[t]{0.24\linewidth}
\centering
\includegraphics[width=1.5in]{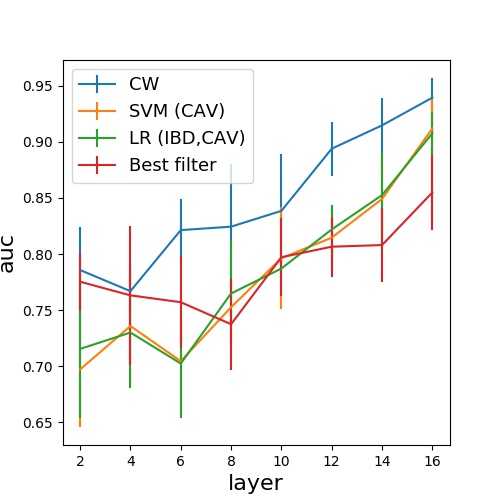}
\end{minipage}%
}
\subfigure[Concept ``book'']{
\begin{minipage}[t]{0.24\linewidth}
\centering
\includegraphics[width=1.5in]{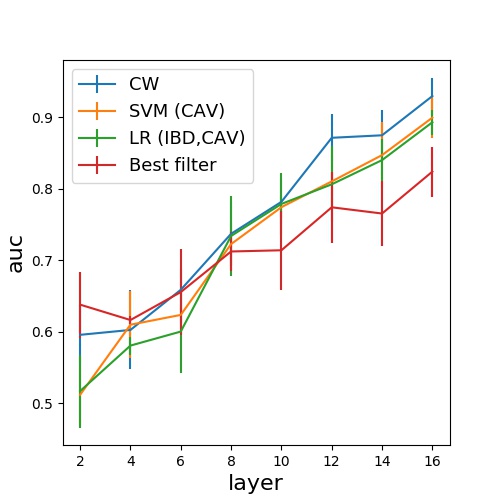}
\end{minipage}%
}
\subfigure[Concept ``bus'']{
\begin{minipage}[t]{0.24\linewidth}
\centering
\includegraphics[width=1.5in]{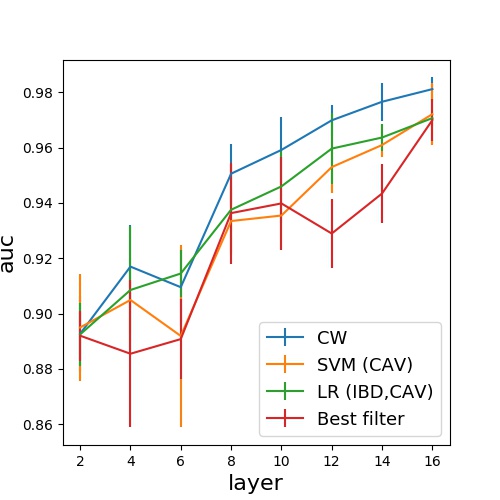}
\end{minipage}%
}
\subfigure[Concept ``car'']{
\begin{minipage}[t]{0.24\linewidth}
\centering
\includegraphics[width=1.5in]{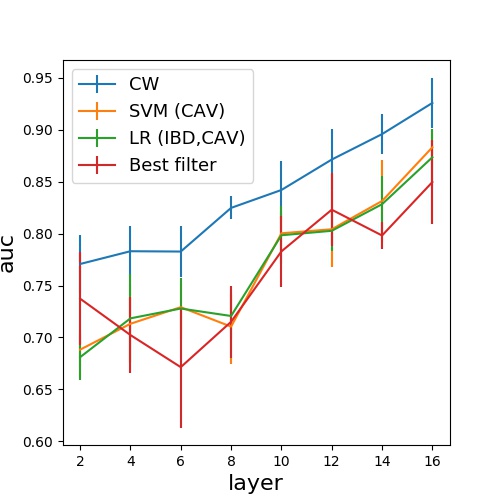}
\end{minipage}%
}
\subfigure[Concept ``dining table'']{
\begin{minipage}[t]{0.24\linewidth}
\centering
\includegraphics[width=1.5in]{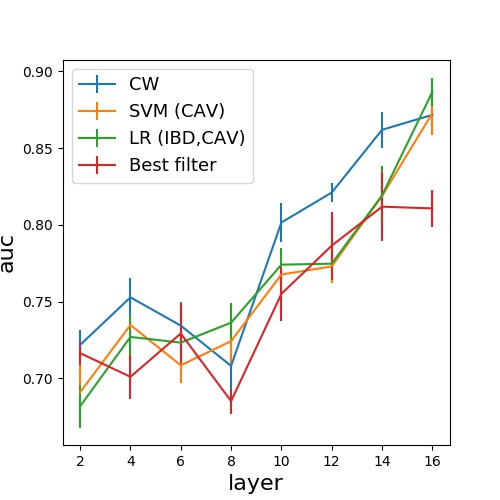}
\end{minipage}%
}
\subfigure[Concept ``horse'']{
\begin{minipage}[t]{0.24\linewidth}
\centering
\includegraphics[width=1.5in]{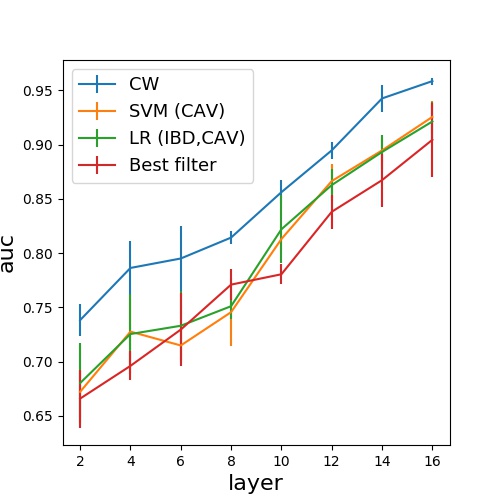}
\end{minipage}%
}
\subfigure[Concept ``person'']{
\begin{minipage}[t]{0.24\linewidth}
\centering
\includegraphics[width=1.5in]{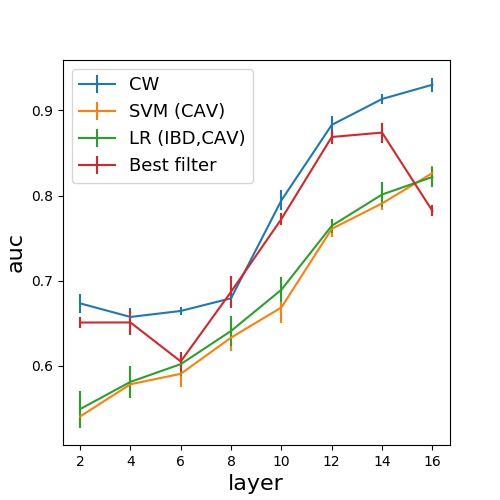}
\end{minipage}%
}
\subfigure[Concept ``potted plant'']{
\begin{minipage}[t]{0.24\linewidth}
\centering
\includegraphics[width=1.5in]{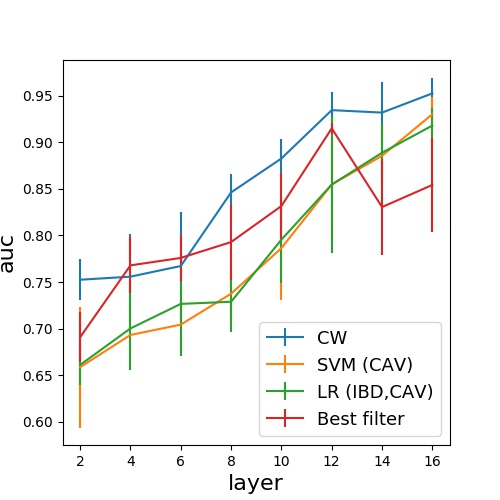}
\end{minipage}%
}
\subfigure[Concept ``sink'']{
\begin{minipage}[t]{0.24\linewidth}
\centering
\includegraphics[width=1.5in]{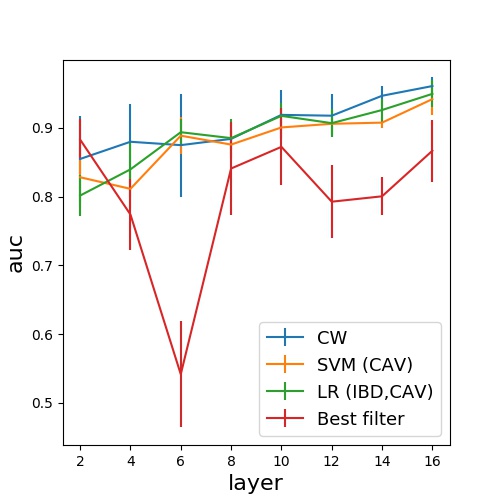}
\end{minipage}%
}
\subfigure[Concept ``umbrella'']{
\begin{minipage}[t]{0.24\linewidth}
\centering
\includegraphics[width=1.5in]{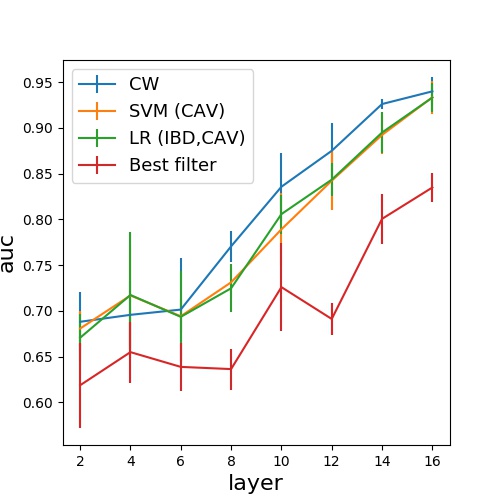}
\end{minipage}%
}
\subfigure[Concept ``wine glass'']{
\begin{minipage}[t]{0.24\linewidth}
\centering
\includegraphics[width=1.5in]{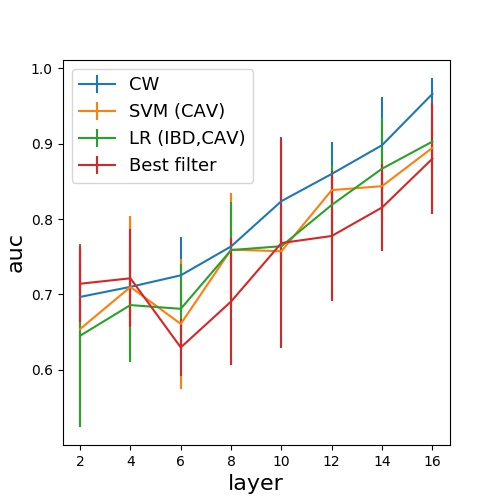}
\end{minipage}%
}
\centering
\caption{Concept purity measured by AUC score on 14 different concepts. Concept purity of CW module is compared to several posthoc methods on different layers. For CW, these figures summarize the results of 16 trained neural networks, each of which had a CW layer containing 7 simultaneous concepts at a different location within the network. For the black box baseline, we use the PlacesCNN neural network \citep{zhou2017places}. The error bar is calculated by the standard deviation over 5 different test sets, and each one is $20\%$ of the entire test set.}
\label{fig:auc_all}
\end{figure}

Supplementary Figure  \ref{fig:auc_all} compares the AUC concept purity of 14 concepts learned by TCAV \cite{kim2018interpretability}, IBD \cite{zhou2018interpreting}, filters in standard CNNs \cite{zhou2014object}, and the CW module in eight different layers ($2^{nd}, 4^{th}, 6^{th}, 8^{th}, 10^{th}, 12^{th}, 14^{th}$, and $16^{th}$ layers). In the figure, the blue line with error bars frequently dominates the AUC across the layers. Therefore, the concepts learned by CW module are generally purer than those learned by other methods. 

\subsection{General Characteristics Concepts}
\label{sec:more_general_concepts}
\begin{figure}[htbp]
    \subfigure[Concepts trained together: cold, moist/damp, warm]{
    \centering
    \begin{minipage}[t]{1\linewidth}
    \includegraphics[width=6.75in]{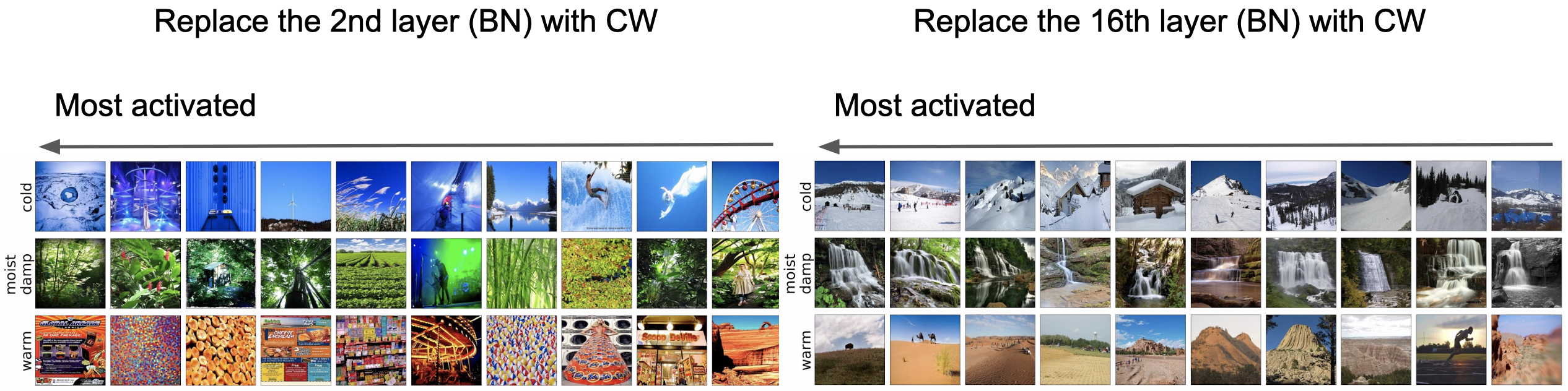}
    \end{minipage}%
    } 
    \subfigure[Concepts trained together: metal, rubber/plastic, wood]{
    \centering
    \begin{minipage}[t]{1\linewidth}
    \includegraphics[width=6.75in]{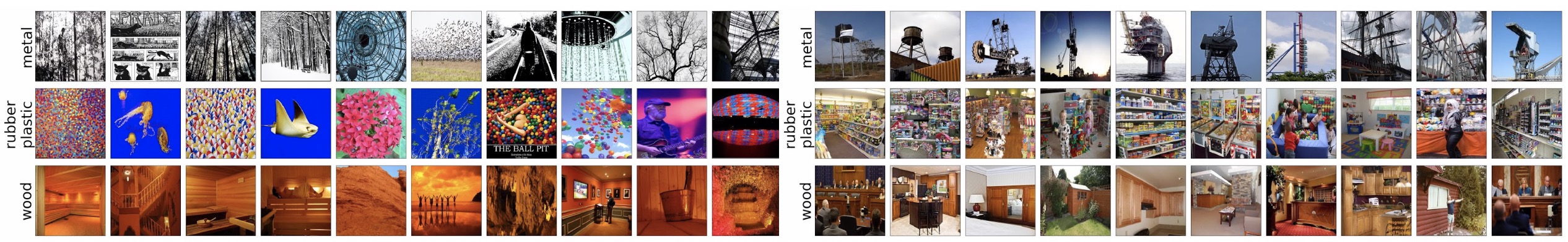}
    \end{minipage}%
    } 
  \caption{Top-10 activated images when CW is trained on concepts that describe general properties of objects and scenes. In the second layer (left subfigures), the concepts seem to be represented by simpler primitive concepts, such as color and texture, whereas in the $16^{th}$ layer (right subfigures) the most activated images seem to correctly capture the high-level meaning of the concept.}
  \label{fig:top10_general}
\end{figure}

The concept bank describing general characteristics of objects and scenes is obtained from the SUN Attribute Database \cite{patterson2012sun}. The attributes in the dataset are used as concepts, and images given three (out of three) MTurk votes on having such attributes are selected to form the concept datasets. Here we train the CW module on two groups of concepts describing weather of the scene (``cold,'' ``moist/damp,'' ``warm''), and materials of objects in the scene (``metal,'' ``rubber/plastic,'' ``wood''). Supplementary Figure  \ref{fig:top10_general} shows the top-10 activated images along these concepts' axes. This figure demonstrates that the CW module can also decently capture high-level meaning (right column) and low-level aspects (left column) of both types of general concepts.

\section{Top Activated Images Visualized with Empirical Receptive Fields}\label{sec:receptive_fields}
\begin{figure}[htbp]

\subfigure[$2^{nd}$ layer]{
    \centering
    \includegraphics[scale=0.44]{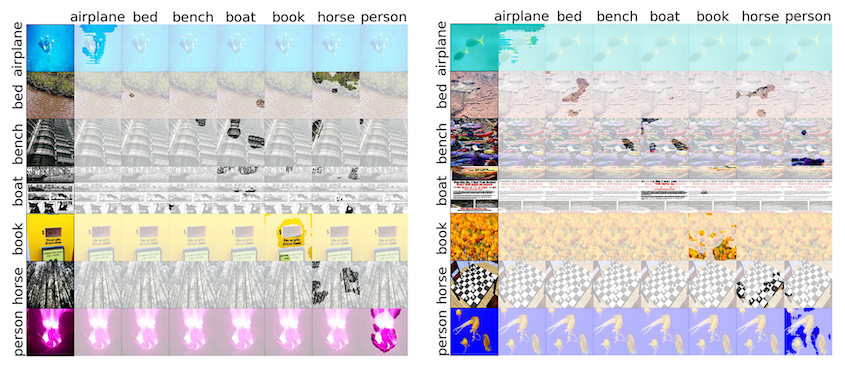}
}
\subfigure[$12^{th}$ layer]{
    \centering
    \includegraphics[scale=0.44]{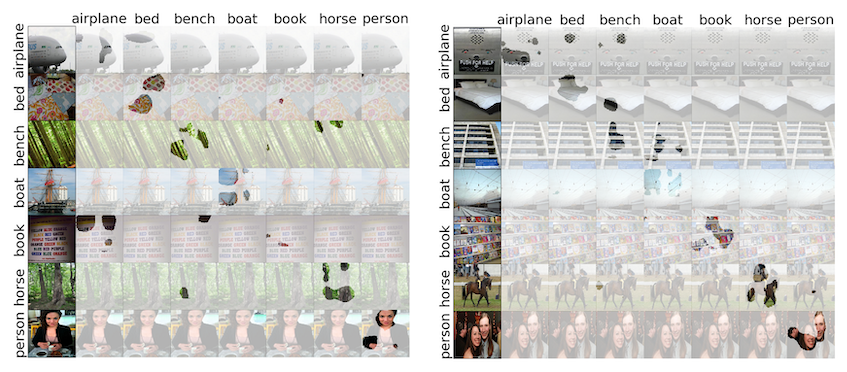}
    }
\subfigure[$16^{th}$ layer]{
    \centering
    \includegraphics[scale=0.44]{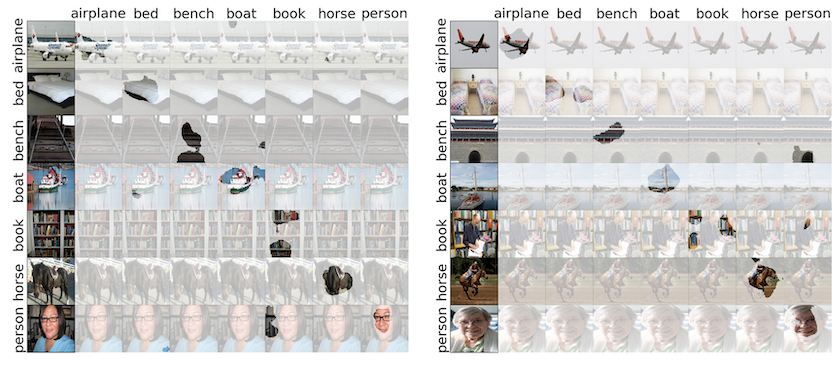}
}
\caption{Some top activated images visualized with empirical receptive fields (highlighted regions). (a) When CW is applied to the $2^{nd}$ layer; (b) when CW is applied to  the $12^{th}$ layer; (c) when CW is applied to  the $16^{th}$ layer. In every subfigure, the leftmost column contains the most activated image for each concept axis. For each image, we calculate its empirical receptive fields on different axes, shown as the 7 images on the right. The empirical receptive field tends to be larger on the portions of the image that are important for recognizing the correct concept.}

\label{fig:empirical_RF}

\end{figure}

To show what local feature could be detected along each concept axis, we visualize the top activated images with the empirical receptive field \cite{zhou2014object}. Empirical receptive fields, in our case, are locations in the image, such that when we black them out, they lead to the greatest reduction in activation values on the different axes of the CW output. We have used $32\times 32$ random covering patches and a stride of $5$ for the sliding window. Supplementary Figure  \ref{fig:empirical_RF} shows the visualization results when CW is applied to the $2^{nd}$, $12^{th}$ and $16^{th}$ layer. Generally, the top activated images for a concept tend to have a larger receptive field on that concept's axis. For an early layer (the $2^{nd}$ layer), the features captured by the concept axes appear to be color and textures. As we proceed to deeper layers, concepts learned by CW become closer to the concepts they aim to represent. For example, in the $12^{th}$ layer, the ``horse'' axis looks at image segments similar to horse legs and the ``person'' axis looks mainly at the hands and faces of people. In the $16^{th}$ layer, the ``horse'' axis is looking at the body of the horse and ``person'' axis is looking at the person's face; both are more representative features. Interestingly, when two concepts occur in the same image, the two concept axes can detect the correct local regions corresponding to these concepts (e.g., the image containing both ``book'' and ``person'' on the $5^{th}$ row of the bottom right subfigure).

\section{Case Study: Skin Lesion Diagnosis}
\label{sec:isic_app}
In this section, we provide a case study of a medical imaging dataset of skin lesions. The dataset of dermoscopic images is collected from the ISIC archive \cite{isic2020}. Because the dermoscopic images corresponding to different diagnoses vary greatly in appearance, we focus on predicting whether a skin lesion is malignant for each of the histopathology images (9058 histopathology images in total). We choose ``age $<$ 20'' and ``size $\geq$ 10 mm'' as the concepts of interest and select the images with corresponding meta information to form the concept datasets. We chose these concepts due to their availability in the ISIC dataset. The cutoff, for instance, of 10mm is used commonly for evaluation of skin lesions \citep{lewis1998}. Details about the experimental results are shown in the following order: test accuracy, separability of latent representation, AUC concept purity, correlation of axes, and concept importance.
\subsection{Test Accuracy}
\label{sec:acc_isic}
We trained both a standard ResNet18 and a ResNet18 with CW on $80\%$ of the dataset and tested it on the other $20\%$. Since the two classes are imbalanced, we measured the balanced accuracy to compare their performances. The test balanced accuracy of standard ResNet18 is $71.65\%$ while ResNet18 with CW achieves $72.26\%$ test balanced accuracy (this is the average over different layers CW was applied to). Thus, adding CW improved performance over the black box; this may have resulted from whitening, which acts as a regularizer. The latent representation in the standard neural network may be elongated due to the inter similarity of the dermoscopic images (empirically shown in Section \ref{sec:correlation_isic}), potentially leading to worse performance. This is why whitening could have provided better numerical conditioning for the gradient, as discussed also by \cite{huang2018decorrelated}.

\subsection{Separability of Latent Representation}
\begin{figure}[t]
    \subfigure[BN module]{
    \centering
    \begin{minipage}[t]{0.45\linewidth}
    \includegraphics[width=2.8in]{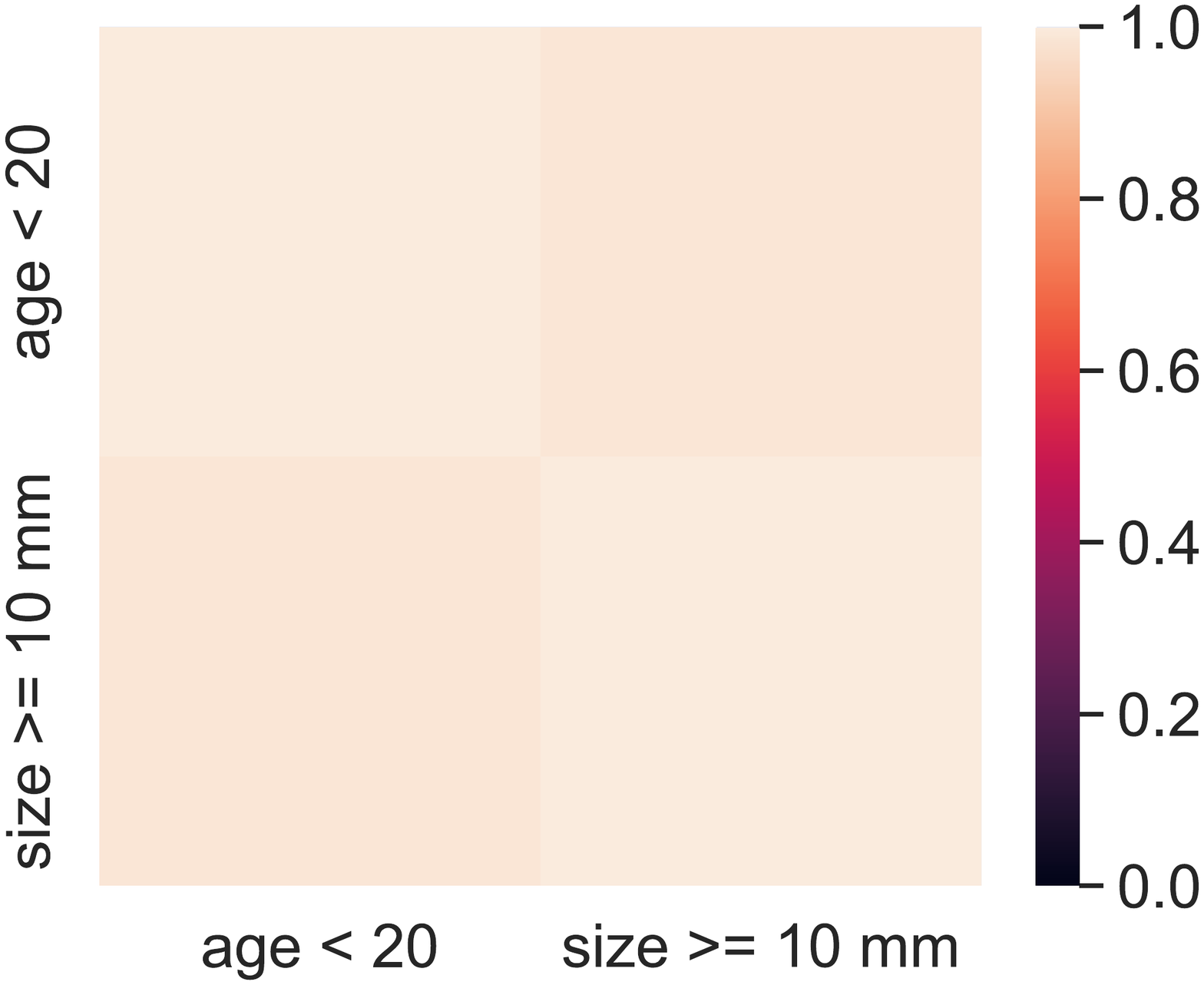}
    \end{minipage}%
    }
    \subfigure[CW module]{
    \centering
    \begin{minipage}[t]{0.45\linewidth}
    \includegraphics[width=2.8in]{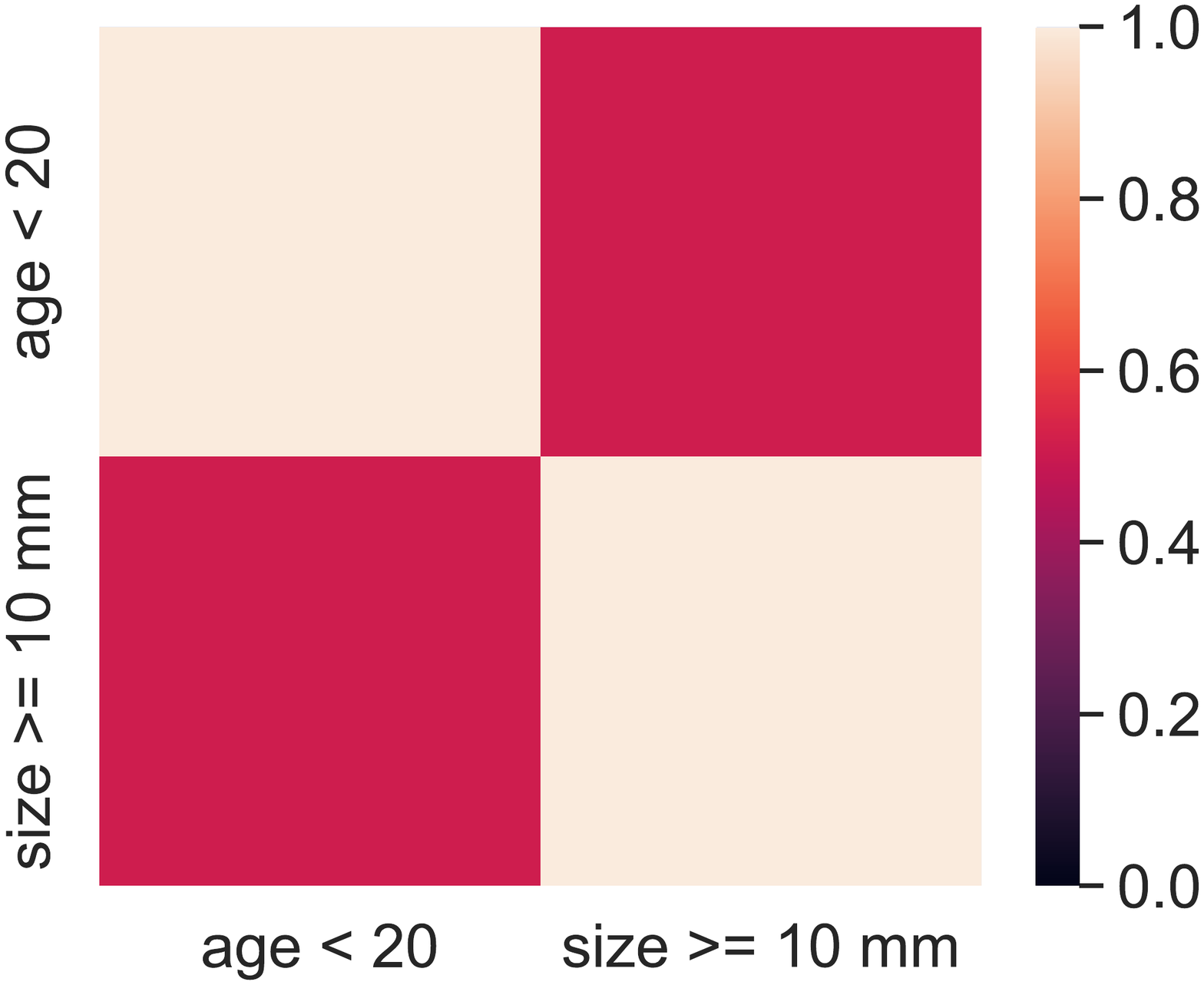}
    \end{minipage}%
    }
  \caption{Normalized intra-concept and inter-concept similarities (ISIC dataset). Diagonal values are normalized average similarities between latent representations of images of the same concept; off-diagonal values are normalized average  similarities between latent representations of images of different concepts. (a) when the $16^{th}$ layer is a BN module; (b) when $16^{th}$ layer is a CW module.}
  \label{fig:inner_product_isic}
\end{figure}

Similar to experiments on the Places dataset, we measured the separability of concepts in the latent space of CW and a standard ResNet (see Figure \ref{fig:inner_product_isic}). When including the CW module, the separability of concepts is also significantly improved.
\label{sec:inter_inner_isic}

\subsection{AUC Concept Purity}
\label{sec:auc_isic}
\begin{figure}[ht]
    \subfigure[Concept ``age $<$ 20'']{
    \centering
    \begin{minipage}[t]{0.45\linewidth}
    \includegraphics[width=2.8in]{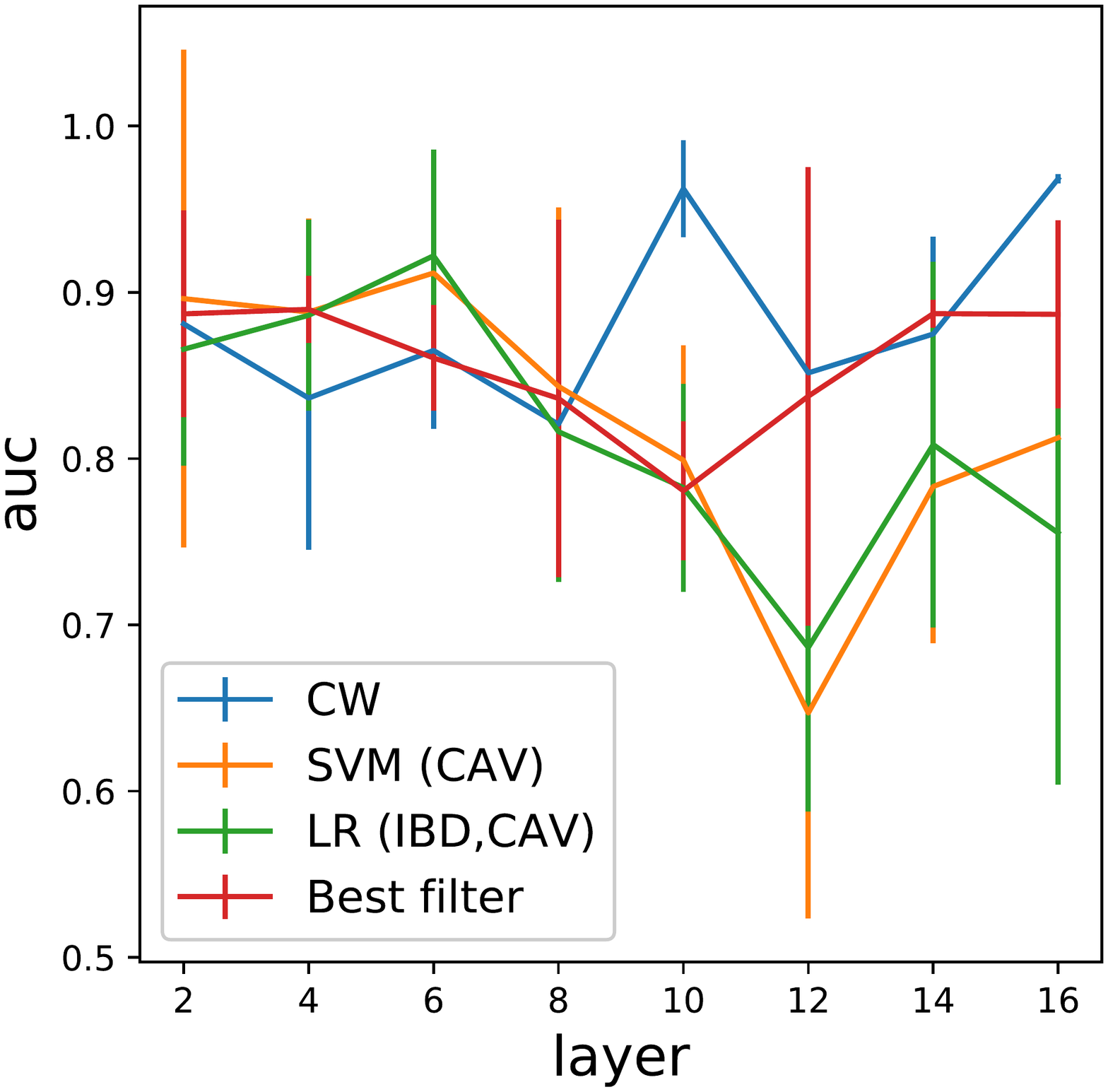}
    \end{minipage}%
    } 
    \subfigure[Concept ``size $\geq$ 10 mm'']{
    \centering
    \begin{minipage}[t]{0.45\linewidth}
    \includegraphics[width=2.8in]{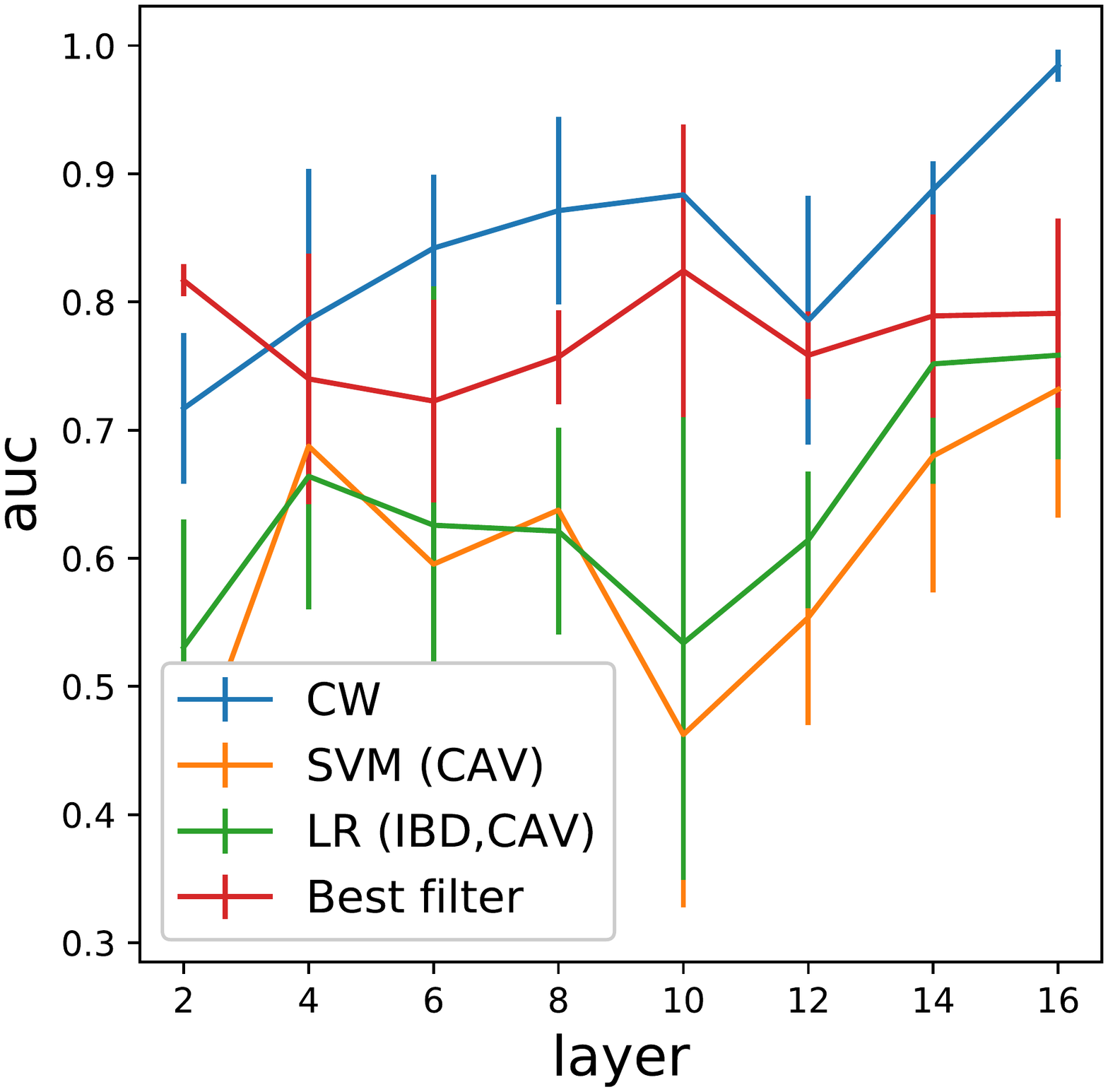}
    \end{minipage}%
    } 
  \caption{Concept purity measured by AUC score (ISIC dataset). Concept purity of CW module is compared to other posthoc methods on different layers. The error bar is the standard deviation over 5 different test sets, and each one is $20\%$ of the entire test set.}
  \label{fig:auc_isic}
\end{figure}

Similar to experiments on the Places dataset, we quantitatively compare the purity of learned concepts with concept-based posthoc methods. As is shown in Figure \ref{fig:auc_isic}, the concept ``age $<$ 20'' is purer using the CW module. All methods were approximately tied in the purity of the concept ``size $\geq$ 10 mm.''

\subsection{Correlation of Axes}
\label{sec:correlation_isic}
\begin{figure}[ht]
    \subfigure[BN module]{
    \centering
    \begin{minipage}[t]{0.45\linewidth}
    \includegraphics[width=2.8in]{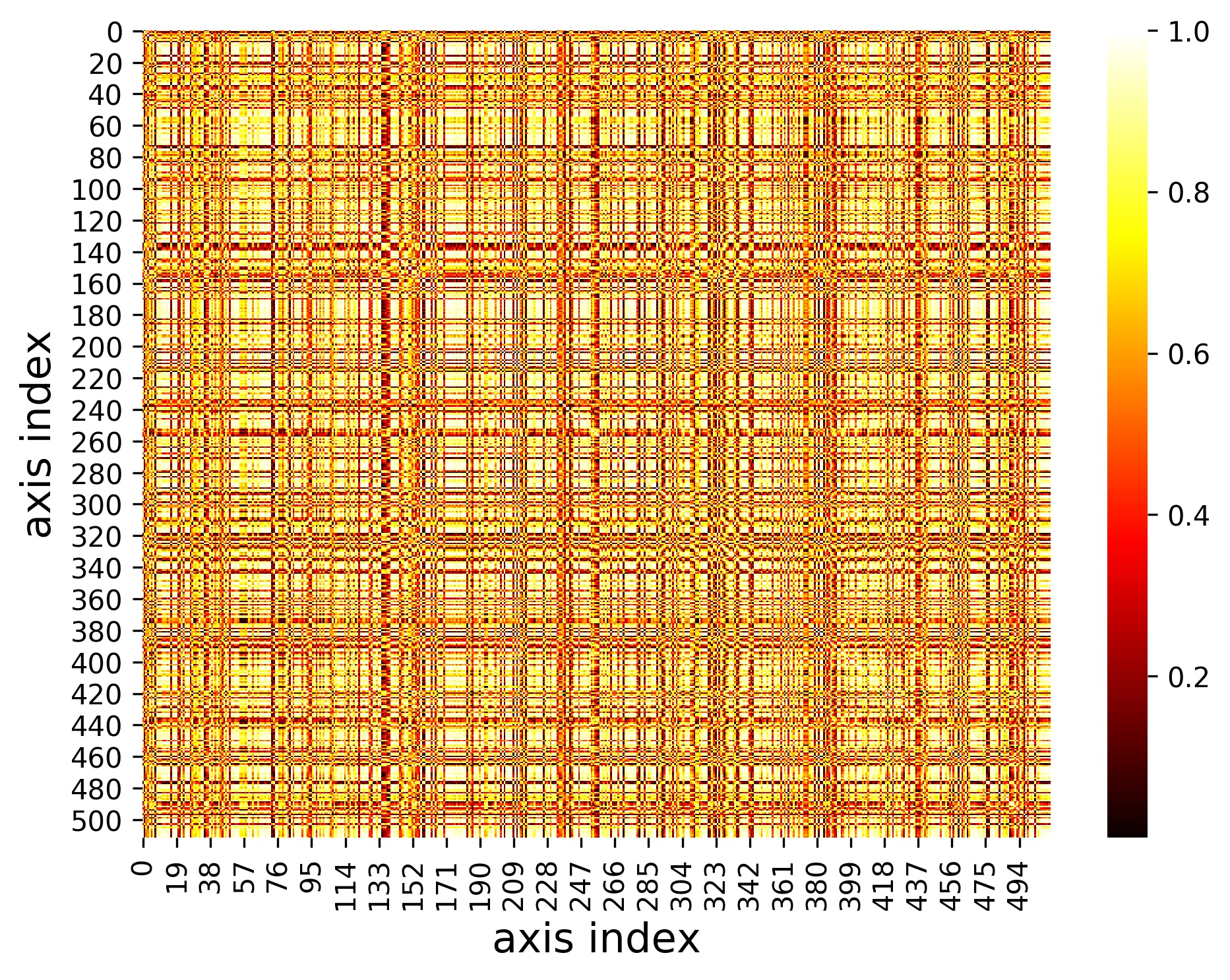}
    \end{minipage}%
    }
    \subfigure[CW module]{
    \centering
    \begin{minipage}[t]{0.45\linewidth}
    \includegraphics[width=2.8in]{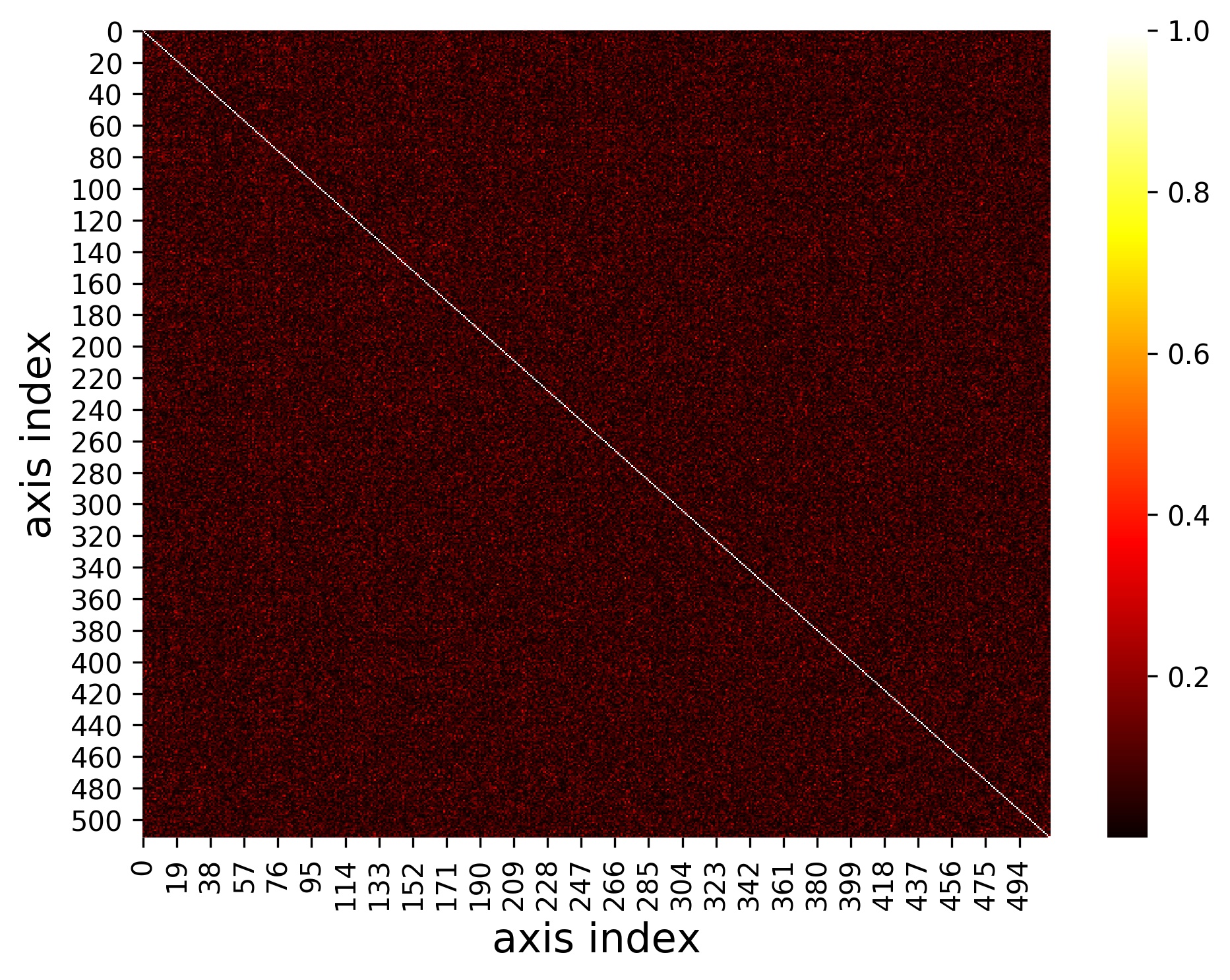}
    \end{minipage}%
    }
  \caption{Absolute correlation coefficient of every feature pair in the $16^{th}$ layer (ISIC dataset). (a) when the $16^{th}$ layer is a BN module; (b) when $16^{th}$ layer is a CW module.}
  \label{fig:correlation_isic}
\end{figure}

Figure \ref{fig:correlation_isic} shows the correlation of axes in the $16^{th}$ layer of ResNet18 with and without the CW module. Shown in Figure \ref{fig:correlation_isic}(a), the correlations of different axes in standard neural networks are very strong (near 1 in many cases). Such highly correlated data distributions in the latent space may negatively influence both the concept separation and stochastic gradient descent, consistent with results in Section \ref{sec:acc_isic} and \ref{sec:inter_inner_isic}. On the contrary, CW can decorrelate the latent space successfully (shown in Figure \ref{fig:correlation_isic}(b)).

\subsection{Concept Importance}
\label{sec:concept_importance_isic}

\begin{figure}[ht]
    \centering
    \includegraphics[width=5in]{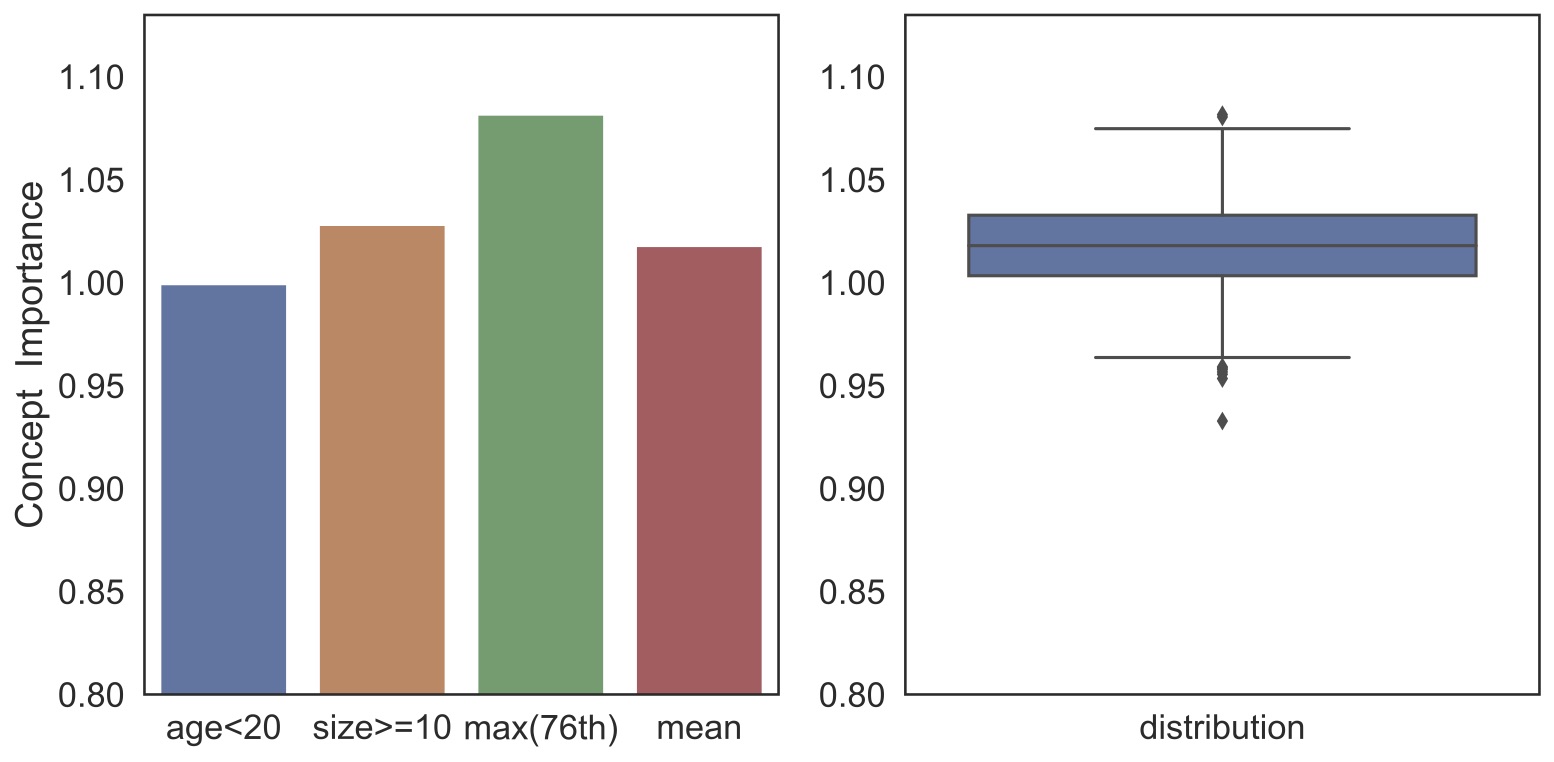}
    \caption{Concept importance measured on each axis when CW is applied to the $16^{th}$ layer (ISIC dataset). The figure on the left shows the concept importance of the axes representing the concepts ``age $<$ 20'' and ``size $\geq$ 10 mm,'' as well as the max and mean concept importance over the set of axes. In order to calculate the latter two quantities, we compute the concept importance of each axis. Then we find the axis with the maximum concept importance (which is the 76th axis) and, for comparison, we calculate the mean of the concept importance values over all the axes. The box plot on the right roughly shows the distribution of concept importance among the 512 axes in the latent space.}
    \label{fig:concept_importance_isic}
\end{figure}

Similar to experiments on the Places dataset, we measure the concept importance scores of concepts in the ISIC dataset. Since the dataset only has two classes, we can measure the contribution to the entire classification problem, using balanced binary cross entropy loss for $e^{(j)}_{\rm switch}$ and $e_{\rm orig}$. Figure \ref{fig:concept_importance_isic} shows the concept importance of different axes of the latent space when CW is applied to the $16^{th}$ layer. We choose the $16^{th}$ layer to investigate because the concepts are purer in the layer as shown in Figure \ref{fig:auc_isic}. We measure the concept importance of the two concepts we selected and the max and mean concept importance of the 512 axes in the latent space (left subplot of Figure \ref{fig:concept_importance_isic}). To compare them with the concept importance of other axes, we also visualize the rough distribution of the concept importance with a box plot (right subplot of Figure \ref{fig:concept_importance_isic}). We observe that the concept ``age $>$ 20'' is not important at all ($\approx 1.0$). The concept ``size $\geq$ 10mm'' is more important than most axes (approximately the third quartile among the 512 axes). This concept is known to be important for the way physicians interpret skin lesions \cite{walter2013using}. 

\begin{figure}[t]
    \centering
    \includegraphics[width=5in]{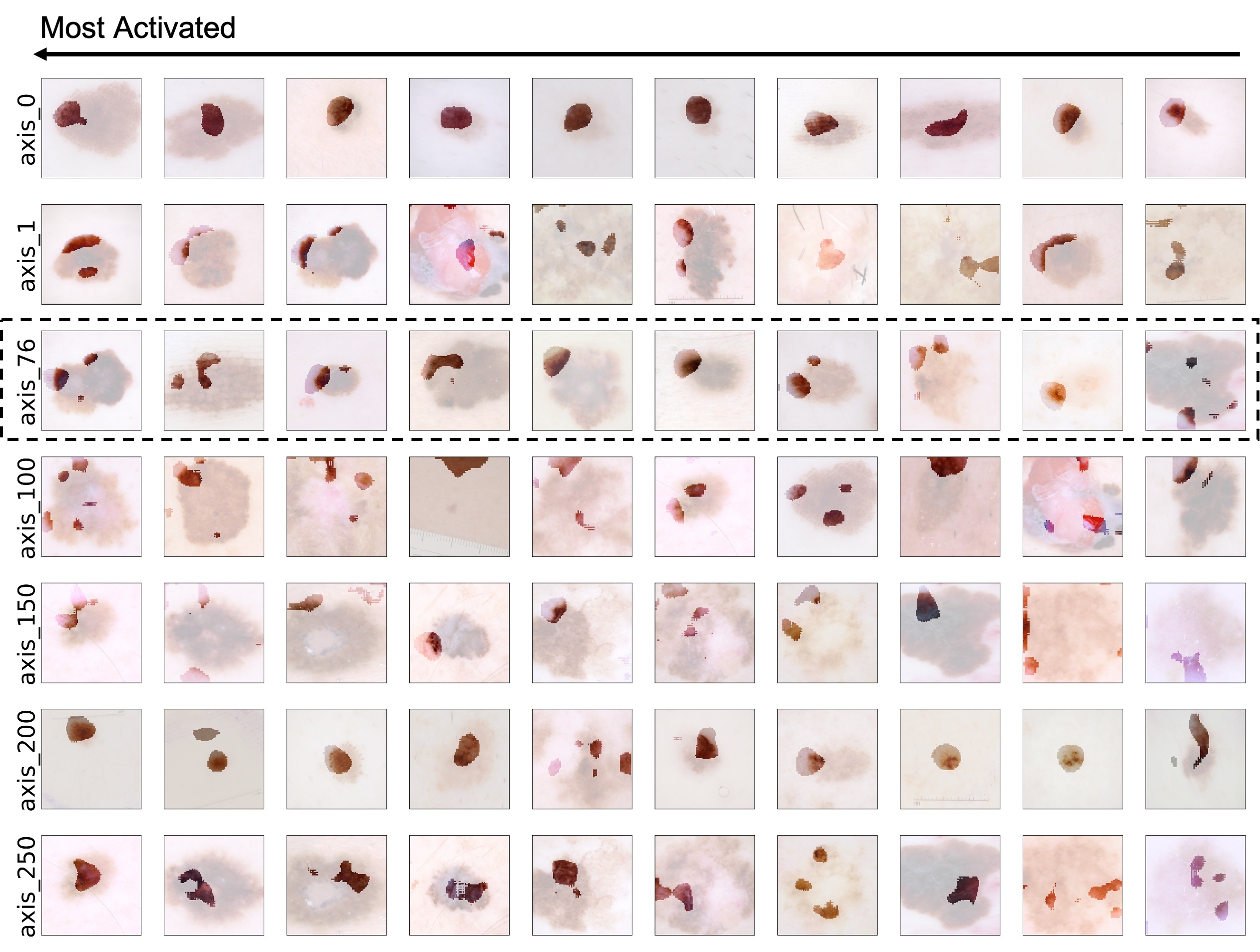}
    \caption{Top 10 activated images on different axes plotted with empirical receptive fields (highlighted region). Axis 76 (most important axis) is highlighted by a dashed box and plotted with other axes (Axis 0, 1, 100, 150, 200 and 250). Axis zero is age, axis one is size, whereas the other axes are not trained as concept axes. Axis 76 seems to more consistently focus on the borders of the lesion, indicating that in future work one might add a concept axis for irregular border.}
    \label{fig:top10_lesion_region}
\end{figure}

It is interesting to contemplate what concept the most important axis (the $76^{th}$ axis) might represent. This axis was not trained to represent a concept, but insight from examining it might lead to possible ideas for concepts we would consider in the future. In Figure \ref{fig:top10_lesion_region}, we visualize the top-10 activated images along this interesting axis, as well as other axes (axes 0, 1, 100, 150, 200, 250) for comparison. We highlight the empirical receptive fields \cite{zhou2014object} on the images. Compared to other axes, the empirical receptive fields of $76^{th}$ axis seems to more consistently focus on the borders of the lesions. The lesion border is well known to be important for early detection of melanoma; an irregular border is a major factor, and is even more important than the overall size of the lesion \cite{walter2013using}. 
This observation naturally leads to a direction for future research: create a concept axis for irregular lesion borders. Since the ISIC dataset does not have each image labeled as to whether the lesion's borders are irregular, this would need to be labeled by a physician in future work. Doing this would allow us to measure the importance of irregular borders for predicting malignancy of skin lesions by a neural network model.

\end{document}